\documentclass[11pt]{article}

\usepackage[preprint]{acl}

\usepackage{times}
\usepackage{latexsym}

\usepackage[T1]{fontenc}

\usepackage[utf8]{inputenc}

\usepackage{microtype}

\usepackage{inconsolata}

\usepackage{graphicx}

\usepackage{subcaption}
\usepackage[most]{tcolorbox}
\usepackage{longtable}
\usepackage{multirow}
\usepackage{colortbl}
\usepackage{tabularx}
\usepackage{makecell}
\usepackage{xcolor}
\usepackage{array}
\usepackage{tikz}
\tcbuselibrary{breakable}
\usepackage{fvextra}
\usepackage{booktabs} 
\usepackage{amsmath}
\usepackage{amssymb}
\usepackage{mathtools}
\usepackage{amsthm}
\usepackage{fontawesome5}
\usepackage{rotating}

\title{Can Scientific Claims Be Removed from Large Language Models? A Systematic Evaluation of Claim-Level Unlearning}

\author{Snigdha Paul \\
  TCS Research, India \\
  \texttt{snigdha.paul@tcs.com} \\\And
  Manasi Patwardhan \\
  TCS Research, India \\
  \texttt{manasi.patwardhan@tcs.com} \\\And
  Arman Cohan \\
  Yale University, USA \\
  \texttt{arman.cohan@yale.edu} \\}

\begin{document}
\maketitle
\begin{abstract}
Language models (LMs) are trained on static scientific corpora, whereas scientific knowledge continuously evolves through correction and revision. Scientific claims encoded within these models may later become retracted, disproven, or updated by subsequent research, creating the risk of disseminating outdated information in scientific workflows. This creates a need for LMs to forget obsolete scientific claims. Machine unlearning offers a promising solution by enabling knowledge removal while maintaining overall model utility. Existing studies primarily investigate instance-level forgetting; however, scientific claims introduce additional challenges because they are interconnected, and continually evolving. To address this gap, we introduce the task of Scientific Claim Unlearning and present a new benchmark, SciUnlearn. We show that current unlearning approaches are unable to effectively eliminate claim-level knowledge and often achieve only superficial suppression, highlighting the need for specialized methods designed for structured knowledge removal.
\end{abstract}

\section{Introduction}

Large Language Models (LLMs) are increasingly used as scientific assistants \cite{Exler2026LLMBasedSA}, supporting hypothesis generation \cite{li-etal-2025-chain-ideas, garikaparthi-etal-2025-iris}, literature reviews \cite{tang-etal-2025-large}, and scientific discovery \cite{Gottweis2025TowardsAA, alphaevolve, Lu2024TheAS}. However, LLMs are trained on static corpora with fixed cutoffs, which conflict with the dynamic and self-correcting nature of scientific knowledge.

LLMs can internalize scientific claims that are later invalidated by retractions or errors \cite{Lesn2006ASA}, allowing false knowledge to persist in tasks like ideation \cite{Alkan2025ASO}, claim verification \cite{pradeep-etal-2021-scientific} etc, also harming in domains like healthcare. LLMs lack mechanisms to deprecate outdated claims, leading to \emph{epistemic inertia} (Example in App.~\ref{sec:motivation_diagram}) despite new contradictory evidence \cite{Caliskan2016SemanticsDA, Gonen2019LipstickOA}. A good example is models treating the fabricated disease “Bixonimania” as real \cite{StokelWalker2026FakeDisease}. Also, some scientific knowledge later becomes harmful, sensitive, or dual‑use \cite{Urbina2022DualUO}, raising regulatory concerns such as the \emph{right to be forgotten} \cite{Zhang2023RightTB}, yet such knowledge is extremely difficult to remove once embedded in LLMs.

These challenges motivate scientific claim unlearning, where models remove outdated claims. Although unlearning has been studied for personal data, copyrighted content and safety‑critical knowledge \cite{Shi2024MUSEMU, Jiang2025LargeLM}, scientific claims remains largely unexplored. Existing work does not address claim-level unlearning or provide systematic benchmarks \cite{Yang2025UnlearningAA}.

Here, we define the task of scientific claim unlearning (\S~\ref{sec:task_def}) and introduce a benchmark to evaluate claim-level forgetting (\S~\ref{sec:dataset}). We adapt baseline unlearning algorithms to this setting (\S~\ref{sec:algo}) and analyze the trade-offs between effective forgetting and scientific knowledge retention (\S~\ref{sec:results}). The benchmark and code are available at \url{https://github.com/snigdhapaul2003/Scientific-Claim-Unlearning/} under Apache License.

\begin{figure*}[ht]
    \centering
    \includegraphics[width=0.80\textwidth]{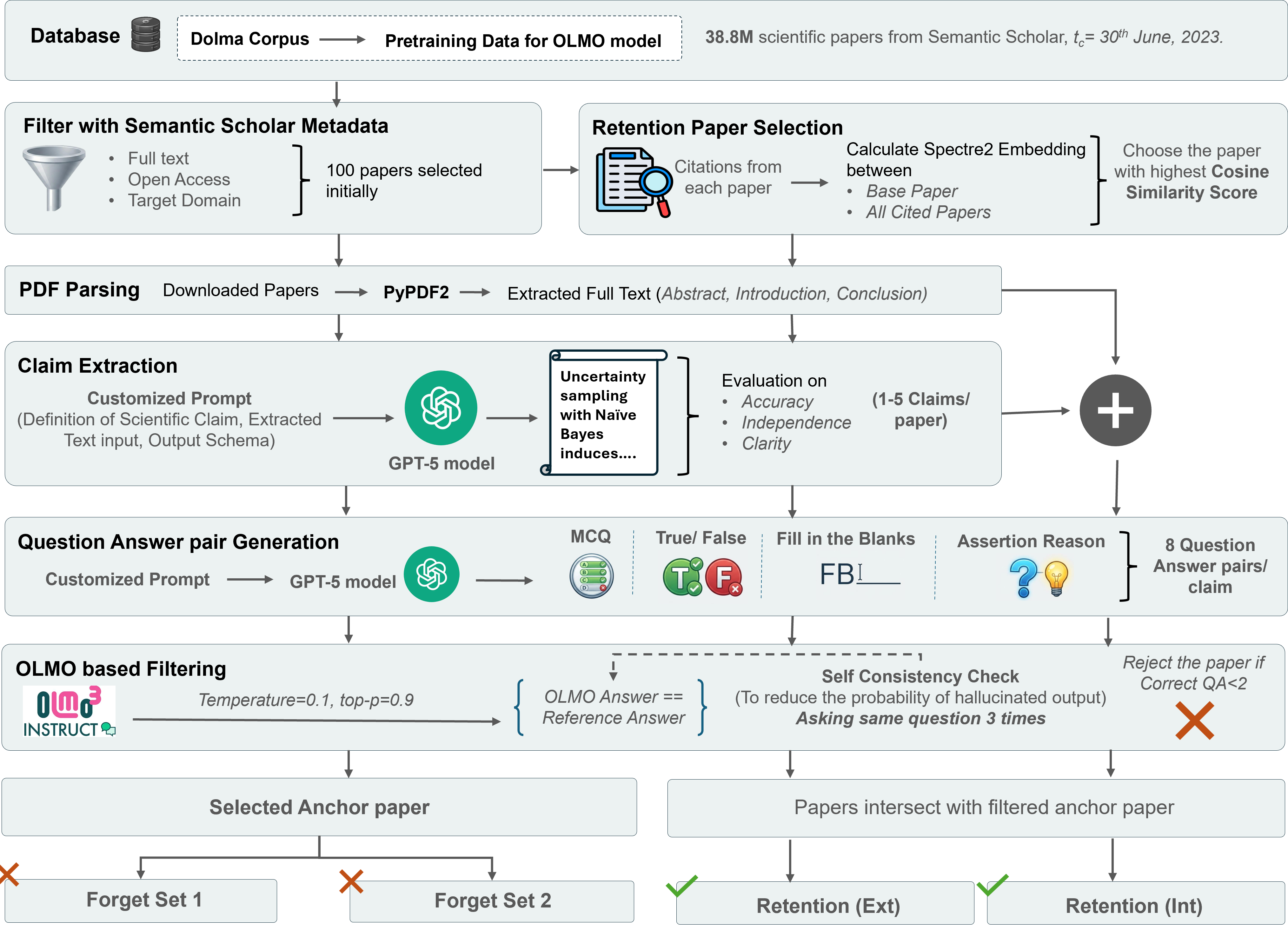}
    \caption{SciUnlearn Dataset generation pipeline. 
    }
    \label{fig:pipeline}
\end{figure*}

\section{Related Works}
\label{sec:related_works}

Scientific claim extraction structures knowledge from scientific text \cite{park-blake-2012-identifying, Wei2023ClaimDistillerSC}, with recent LLM pipelines combining extraction, retrieval, and verification \cite{10.1145/3701716.3717752}. Existing datasets and tracking systems cover multiple domains and temporal claim evolution \cite{achakulvisut2019claim, wadden-etal-2020-fact, diggelmann2020climatefever, wang2025sciencemeter, Pramanick2026ClaimFlowTT}, but focus on organizing claims rather than unlearning invalid ones. Machine unlearning removes targeted training influence while preserving utility \cite{Bourtoule2019MachineU}, and has been extended to LLMs for privacy, copyright, and harmful-knowledge removal \cite{Liu2024RethinkingMU, tofu2024, Shi2024MUSEMU, Li2024TheWB, jin2024rwku}. Methods include gradient-based updates, preference optimization and parameter-efficient tuning\cite{jin-etal-2025-unlearning, Zhang2024NegativePO, Liu2025LUNEEL}; however, removed knowledge may persist in representations and re-emerge across contexts \cite{Jia2025TheEI, rybak2026rebelhiddenknowledgerecovery}.

\section{Scientific Claim Unlearning Task Definition}
\label{sec:task_def}

Let $\mathcal{M}_{\theta}$ denote a pretrained language model with parameters $\theta$. We assume availability of dataset $\mathcal{D} = \{\mathcal{F}_1, \mathcal{F}_2, \mathcal{R}_{ext}, \mathcal{R}_{int}\}$, where  $\mathcal{F}_1 \cap \mathcal{F}_2 = \emptyset$, are forget sets with distinct types of paraphrased question answer pairs $(q,a)$ derived from a shared set of underlying scientific claims $\mathcal{S}$, extracted from scientific papers $\mathcal{P}$ included in the pre-training corpus of $\mathcal{M}_{\theta}$. This enables evaluation of \textit{claim level unlearning}: when a model is unlearned on $\mathcal{F}_1$, it is expected to also forget $\mathcal{F}_2$, and vice-versa, thereby testing whether the underlying claim has been removed rather than only specific superficial forms. The retain sets 
are designed to validate the preservation of non-target knowledge such that, 
$(\mathcal{R}_{ext} \cup \mathcal{R}_{int}) \cap (\mathcal{F}_1 \cup \mathcal{F}_2) = \emptyset$.
The external retain set $\mathcal{R}_{ext}$ consists of cited prior knowledge from papers referenced by the anchor papers $\mathcal{P}$, while the internal retain set $\mathcal{R}_{int}$ contain non claim based contextual information from the same paper.

Given the base model $\mathcal{M}_{\theta}$ and a forget set $\mathcal{F}_i$ ($i = 1$ or $2$), an unlearning algorithm $\mathcal{A}$ updates the model parameters as $\theta' = \theta + \mathcal{A}(\theta, \mathcal{F}_i, \mathcal{R})$ and $\mathcal{R} \subseteq \{\mathcal{R}_{ext}, \mathcal{R}_{int}\}$. Utilization of $\mathcal{R}$ is optional depending on the algorithm $\mathcal{A}$. Building upon the standard formulation of LLM unlearning \cite{Liu2024RethinkingMU}, the objective is defined by the Eq. \ref{eq:unlearn_obj}.

{
\small
\begin{equation}
\min_{\theta'} \;
\underbrace{\mathbb{E}_{(q,a)\in \mathcal{F}_i} \left[ \ell_f(a|q;\theta') \right]}_{\text{Forget}}
+
\underbrace{\lambda \, \mathbb{E}_{(q,a)\in \mathcal{R}} \left[ \ell_r(a|q;\theta') \right]}_{\text{Retain}}
\label{eq:unlearn_obj}
\end{equation}
}

\begin{figure*}[ht]
    \centering
    \includegraphics[width=0.95\textwidth]{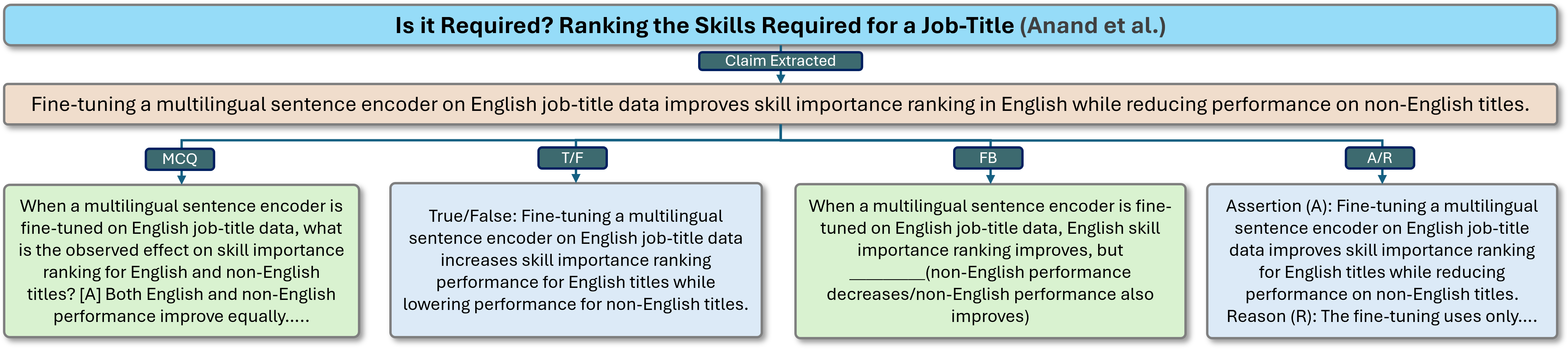}
    \caption{Example claim and QA in SciUnlearn dataset.}
    \label{fig:Benchmark_sample}
\end{figure*}

where $\ell_f(a|q;\theta') = \log p_{\theta'}(a|q)$ denotes the forgetting loss, and $\ell_r(a|q;\theta') = - \log p_{\theta'}(a|q)$ denotes the retention loss. $\lambda \geq 0$ balances forgetting and retention. Performance is measured on a disjoint forget set as  
$\mathbb{E}_{(q,a)\in \mathcal{F}_j}\!\left[\ell_f(a\mid q;\theta')\right], \; j \neq i.$, which evaluates claim-level forgetting beyond superficial forgetting.

\section{SciUnlearn Dataset}
\label{sec:dataset}

To study scientific claim unlearning, we build three benchmark datasets from the Dolma corpus ($\mathcal{C}$) \cite{soldaini-etal-2024-dolma}, which includes academic papers from Semantic Scholar \cite{Kinney2023TheSS}. Dolma's scale, recency, and use in training OLMO models ensure that the targeted claims are likely embedded in pretrained models, enabling meaningful evaluation. We further validate this assumption using likelihood-based membership signals and answer-order perturbations (App.~\ref{sec:memorization_reasoning}). The benchmark consists of a computer science dataset $\mathcal{D}_{\text{cs}}$, a medical dataset $\mathcal{D}_{\text{med}}$, and a retracted-paper dataset $\mathcal{D}_{\text{retracted}}$. The first two datasets cover domain-specific claim unlearning in active scientific fields, while $\mathcal{D}_{\text{retracted}}$ contains Dolma papers retracted after 2024 and identified using the Retraction Watch database \cite{retractionwatch2024}. The full pipeline is shown in Fig.~\ref{fig:pipeline}.

We sample 100 recent (2018–2022) open‑access computer science papers from the Dolma corpus using the Semantic Scholar API \cite{Kinney2023TheSS}, forming $\mathcal{C}_f \subset \mathcal{C}$. For each paper $p \in \mathcal{C}_f$, we extract $1$–$5$ claims $\mathcal{S}(p)$ using GPT‑5\footnotemark[1] from the abstract, introduction, and conclusion\footnotemark[2]. We use an \textit{LLM as a judge} to evaluate the extracted claims from all forget and retain set papers based on accuracy, independence, and clarity\footnotemark[2] on a 0–5 scale, using GPT‑5.4, GPT-5.1 and Gemini-3.5-Flash\footnotemark[1]. Table~\ref{tab:claim_quality_consensus} reports the percentage of cases in which at least two of the three models assign a score of 4 or higher out of 5. To validate LLM judgments, we correlate LLM scores with expert annotations on 75 claims from 50 randomly selected papers (App.~\ref{sec:claim_samples}). Among these 50 papers, 24 are from computer science, 13 from medical science, and 13 from the retracted-paper set. We report Gwet's AC2 instead of Cohen's kappa because Cohen's kappa is susceptible to the kappa paradox under highly imbalanced ratings. As shown in Table~\ref{tab:human_llm_ac2}, Gwet's AC2 indicates high human--LLM agreement across accuracy, independence, and clarity, supporting LLM‑based evaluation as a scalable proxy for human scoring.

\begin{table}[ht]
\centering
\footnotesize
\setlength{\tabcolsep}{5pt}
\renewcommand{\arraystretch}{0.9}
\begin{tabular}{lccc}
\toprule
Domain & Accuracy & Independence & Clarity \\
\midrule
CS & 95.31\% & 99.71\% & 99.71\% \\
Medical & 95.03\% & 98.01\% & 99.67\% \\
Retracted & 92.57\% & 96.28\% & 99.32\% \\
\bottomrule
\end{tabular}
\caption{Multi-LLM consensus rates for claim quality evaluation (\% of claims with majority score $\geq 4$).}
\label{tab:claim_quality_consensus}
\end{table}

\begin{table}[ht]
\centering
\footnotesize
\setlength{\tabcolsep}{4pt}
\renewcommand{\arraystretch}{0.9}
\resizebox{\columnwidth}{!}{%
\begin{tabular}{lccc}
\toprule
Criterion & GPT-5.4 & GPT-5.1 & Gemini 3.5 Flash \\
\midrule
Accuracy & 0.90 & 0.91 & 0.84 \\
Independence & 0.88 & 0.86 & 0.87 \\
Clarity & 0.91 & 0.90 & 0.88 \\
\bottomrule
\end{tabular}
}
\caption{Human--LLM agreement (Gwet's AC2)}
\label{tab:human_llm_ac2}
\end{table}

\footnotetext[1]{Models are accessed via API with temperature = 1.0, top‑p = 0.9, and a maximum input length of 40{,}000 characters.}

\footnotetext[2]{The detailed prompts used are given in https://github.com/snigdhapaul2003/Scientific-Claim-Unlearning/prompts.txt.}

For each claim $s \in \mathcal{S}(p)$, we generate eight $(q,a)$ pairs—two each for MCQ, true/false(T/F), fill-in-the-blank(FB), and assertion--reason(AR)\footnotemark[2]. QA pairs are validated using OLMO-3-7B-Instruct ($\mathcal{M}$) with self-consistency checks \cite{NEURIPS2024_b1f78dfc, Xie2025RevealAR}. Responses are evaluated using exact match, ROUGE \cite{lin-2004-rouge}, and semantic similarity \cite{reimers-gurevych-2019-sentence}, with a threshold of 0.7 for non-exact matches. Claims with fewer than two valid question types are discarded. QA pairs are splitted into $\mathcal{F}_1$ and $\mathcal{F}_2$, as discussed in \S~\ref{sec:task_def}. The statistical details of all subsets are presented in Table~\ref{tab:dataset_stats}

\begin{table}[ht]
\centering
\footnotesize
\setlength{\tabcolsep}{3pt}
\renewcommand{\arraystretch}{0.9}
\begin{tabular}{lcccccc}
\toprule
Dataset & Papers & Claims & F1 & F2 & R$_{ext}$ & R$_{int}$ \\
\midrule
$\mathcal{D}_{\text{cs}}$ & 76 & 170 & 536 & 536 & 1060 & 418 \\
$\mathcal{D}_{\text{med}}$ & 78 & 127 & 508 & 508 & 908 & 543 \\
$\mathcal{D}_{\text{retracted}}$ & 91 & 141 & 390 & 390 & 884 & 671 \\
\bottomrule
\end{tabular}
\caption{Overview of the SciUnlearn datasets.}
\label{tab:dataset_stats}
\end{table}

\begin{table*}[!t]
\centering
\tiny
\setlength{\tabcolsep}{3pt}
\renewcommand{\arraystretch}{0.70}
\resizebox{\textwidth}{!}{%
\begin{tabular}{clccccccccccc}
\toprule
\textbf{Model}
& \textbf{Method}
& \multicolumn{2}{c}{\textbf{Forget Set 1 $\downarrow$}}
& \multicolumn{2}{c}{\textbf{Forget Set 2 $\downarrow$}}
& \multicolumn{2}{c}{\textbf{Retain (Ext) $\uparrow$}}
& \multicolumn{2}{c}{\textbf{Retain (Int) $\uparrow$}}
& \multicolumn{3}{c}{\textbf{General Benchmarks}} \\
\cmidrule(lr){3-4}
\cmidrule(lr){5-6}
\cmidrule(lr){7-8}
\cmidrule(lr){9-10}
\cmidrule(lr){11-13}
& & R-F1 & EM & R-F1 & EM & R-F1 & EM & R-F1 & EM & MMLU & Arc-C & HellaSwag \\
\midrule
\multirow{5}{*}{\rotatebox[origin=c]{90}{OLMO}}
& Base Model & 97.54 & 94.21 & 93.92 & 89.17 & 95.40 & 90.75 & 92.44 & 83.49 & 58.75 & 54.35 & 76.02 \\
\cmidrule(lr){2-13}
& GD ($\mathcal{F}_1$) & 71.70 & 58.20 & 93.51 & 88.24 & 93.66 & 84.15 & 90.53 & 72.00 & 56.44 & 50.08 & 70.13 \\
& NPO+RT ($\mathcal{F}_1$) & 88.81 & 81.52 & 93.67 & 89.55 & 92.44 & 86.13 & 90.66 & 82.05 & 58.28 & 49.91 & 69.89 \\
\cmidrule(lr){2-13}
& GD ($\mathcal{F}_2$) & 91.65 & 78.35 & 89.60 & 83.58 & 93.27 & 84.52 & 91.02 & 77.03 & 57.54 & 49.57 & 70.93 \\
& NPO+RT ($\mathcal{F}_2$) & 96.14 & 89.92 & 86.78 & 82.27 & 91.47 & 85.18 & 91.60 & 81.81 & 57.39 & 50.51 & 73.02 \\
\midrule
\multirow{5}{*}{\rotatebox[origin=c]{90}{LLAMA}}
& Base Model & 96.61 & 93.65 & 87.37 & 77.61 & 91.86 & 86.03 & 90.07 & 83.25 & 64.74 & 75.58 & 55.88 \\
\cmidrule(lr){2-13}
& GD ($\mathcal{F}_1$) & 78.33 & 74.06 & 88.32 & 80.03 & 85.45 & 80.37 & 89.00 & 83.25 & 63.01 & 73.91 & 46.67 \\
& NPO+RT ($\mathcal{F}_1$) & 73.56 & 70.33 & 90.91 & 86.38 & 86.53 & 83.01 & 90.35 & 83.73 & 63.35 & 75.28 & 53.41 \\
\cmidrule(lr){2-13}
& GD ($\mathcal{F}_2$) & 93.31 & 88.80 & 59.18 & 37.87 & 76.78 & 65.47 & 81.31 & 73.20 & 60.93 & 74.30 & 47.69 \\
& NPO+RT ($\mathcal{F}_2$) & 96.08 & 93.09 & 78.59 & 66.23 & 87.36 & 78.86 & 85.14 & 77.27 & 64.44 & 75.01 & 55.37 \\
\bottomrule
\end{tabular}
}
\caption{Results of LoRA unlearning using \textbf{$\mathcal{F}_1$} and \textbf{$\mathcal{F}_2$} for GD and NPO+RT on $\mathcal{D}_{\text{cs}}$. $\downarrow$ indicates lower is better (forgetting), while $\uparrow$ indicates higher is better (retention and generalization). OLMo-3-7B-Instruct is referred as OLMO and LLAMA3-8B-Instruct is referred as LLAMA. Additional method results are provided in App.~\ref{sec:additional_lora_results}.}
\label{tab:main_results_merged}
\end{table*}

\begin{table*}[!t]
\centering
\tiny
\setlength{\tabcolsep}{3pt}
\renewcommand{\arraystretch}{0.70}
\resizebox{\textwidth}{!}{%
\begin{tabular}{clccccccccccc}
\toprule
\textbf{Model}
& \textbf{Method}
& \multicolumn{2}{c}{\textbf{Forget Set 1 $\downarrow$}}
& \multicolumn{2}{c}{\textbf{Forget Set 2 $\downarrow$}}
& \multicolumn{2}{c}{\textbf{Retain (Ext) $\uparrow$}}
& \multicolumn{2}{c}{\textbf{Retain (Int) $\uparrow$}}
& \multicolumn{3}{c}{\textbf{General Benchmarks}} \\
\cmidrule(lr){3-4}
\cmidrule(lr){5-6}
\cmidrule(lr){7-8}
\cmidrule(lr){9-10}
\cmidrule(lr){11-13}
& & R-F1 & EM & R-F1 & EM & R-F1 & EM & R-F1 & EM & MMLU & Arc-C & HellaSwag \\
\midrule
\multirow{5}{*}{\rotatebox[origin=c]{90}{OLMO}}
& Base Model & 92.16 & 78.59 & 84.59 & 65.96 & 87.74 & 71.14 & 92.49 & 81.95 & 58.76 & 54.35 & 76.02 \\
\cmidrule(lr){2-13}
& GD ($\mathcal{F}_1$) & 58.20 & 35.78 & 75.40 & 61.05 & 86.78 & 73.78 & 92.77 & 84.34 & 57.24 & 52.47 & 71.07 \\
& NPO+RT ($\mathcal{F}_1$) & 83.09 & 67.36 & 83.93 & 68.07 & 89.69 & 74.77 & 92.49 & 85.63 & 59.50 & 52.73 & 75.30 \\
\cmidrule(lr){2-13}
& GD ($\mathcal{F}_2$) & 83.80 & 68.77 & 55.59 & 32.28 & 88.97 & 74.77 & 93.95 & 86.37 & 56.43 & 51.70 & 72.50 \\
& NPO+RT ($\mathcal{F}_2$) & 92.52 & 85.64 & 73.89 & 81.53 & 91.96 & 85.06 & 94.79 & 88.07 & 59.14 & 53.41 & 73.54 \\
\midrule
\multirow{5}{*}{\rotatebox[origin=c]{90}{LLAMA}}
& Base Model & 86.23 & 71.92 & 76.64 & 56.49 & 80.39 & 63.32 & 90.24 & 82.68 & 64.75 & 75.58 & 55.88 \\
\cmidrule(lr){2-13}
& GD ($\mathcal{F}_1$) & 70.08 & 55.78 & 73.70 & 56.14 & 81.61 & 65.96 & 93.81 & 88.21 & 63.87 & 75.39 & 52.99 \\
& NPO+RT ($\mathcal{F}_1$) & 41.70 & 26.66 & 41.70 & 15.43 & 87.18 & 70.15 & 97.83 & 94.47 & 47.29 & 74.97 & 54.18 \\
\cmidrule(lr){2-13}
& GD ($\mathcal{F}_2$) & 83.32 & 68.77 & 68.34 & 48.77 & 82.94 & 67.62 & 93.90 & 88.39 & 63.10 & 75.29 & 52.82 \\
& NPO+RT ($\mathcal{F}_2$) & 45.98 & 30.18 & 32.79 & 7.37 & 88.86 & 79.07 & 96.06 & 93.00 & 43.20 & 74.48 & 49.57 \\
\bottomrule
\end{tabular}
}
\caption{Results of full parameter unlearning using \textbf{$\mathcal{F}_1$} and \textbf{$\mathcal{F}_2$} for GD and NPO+RT on $\mathcal{D}_{\text{med}}$. $\downarrow$ indicates lower is better (forgetting), while $\uparrow$ indicates higher is better (retention and generalization). OLMo-3-7B-Instruct is referred as OLMO and LLAMA3-8B-Instruct is referred as LLAMA.}
\label{tab:main_results_merged_med}
\end{table*}

For $\mathcal{R}_{ext}$, we select for each $p \in \mathcal{P}$, the most semantically similar cited paper using SPECTER2 embeddings \cite{singh-etal-2023-scirepeval}, extracting claims and QA pairs from each $p$. $\mathcal{R}_{int}$ is built from the same $p$ using non–claim-specific, contextual QA pairs. We ensure no overlap between forget and retain sets (App.~\ref{sec:overlap}). Fig.~\ref{fig:Benchmark_sample} shows an example, with full QA example provided in App.~\ref{sec:sample_qa_eval}. Final statistics of SciUnlearn dataset are presented in Table~\ref{tab:dataset_stats}. Detailed analysis on the Dataset is provided in App.~\ref{sec:dataset_statistics}.

\section{Experimental Set-up }
\label{sec:algo}

We use OLMo-3-7B-Instruct \cite{groeneveld-etal-2024-olmo} for full-parameter and rank-8 LoRA training on a single 80 GB A100 GPU, and extended the experimentation with LLAMA-3-8B-Instruct. We benchmark representative approaches, including Gradient Difference (GD) \cite{Neel2020DescenttoDeleteGM}, Negative Preference Optimization (NPO) \cite{Zhang2024NegativePO}, NPO with retain objective (NPO+RT) \cite{bronec-helcl-2025-atyaephyra}, SimNPO \cite{Fan2024SimplicityPR}, and SimNPO with retain set. All methods are applied to the pretrained model using forget sets $\mathcal{F}_1$ or $\mathcal{F}_2$. Implementation details and hyperparameters are provided in App.~\ref{sec:implementation} and App.~\ref{sec:hyperpaarmeter}. Task-specific performance is evaluated using Exact Match (EM) and ROUGE-L F1 (R-F1), capturing strict and paraphrase-tolerant correctness, respectively. General knowledge retention is assessed on MMLU, ARC-Challenge (AC), and HellaSwag (HS) \cite{Hendrycks2020MeasuringMM, Clark2018ThinkYH, zellers-etal-2019-hellaswag}, ensuring unlearning does not degrade overall language understanding.

\begin{table*}[!t]
\centering
\tiny
\setlength{\tabcolsep}{3pt}
\renewcommand{\arraystretch}{0.70}
\resizebox{\textwidth}{!}{%
\begin{tabular}{clccccccccccc}
\toprule
\textbf{Model}
& \textbf{Method}
& \multicolumn{2}{c}{\textbf{Forget Set 1 $\downarrow$}}
& \multicolumn{2}{c}{\textbf{Forget Set 2 $\downarrow$}}
& \multicolumn{2}{c}{\textbf{Retain (Ext) $\uparrow$}}
& \multicolumn{2}{c}{\textbf{Retain (Int) $\uparrow$}}
& \multicolumn{3}{c}{\textbf{General Benchmarks}} \\
\cmidrule(lr){3-4}
\cmidrule(lr){5-6}
\cmidrule(lr){7-8}
\cmidrule(lr){9-10}
\cmidrule(lr){11-13}
& & R-F1 & EM & R-F1 & EM & R-F1 & EM & R-F1 & EM & MMLU & Arc-C & HellaSwag \\
\midrule
\multirow{5}{*}{\rotatebox[origin=c]{90}{OLMO}}
& Base Model & 94.65 & 87.94 & 93.05 & 84.61 & 94.12 & 87.33 & 94.54 & 87.92 & 58.75 & 54.35 & 76.02 \\
\cmidrule(lr){2-13}
& GD ($\mathcal{F}_1$) & 65.49 & 44.61 & 84.96 & 78.46 & 91.72 & 86.31 & 95.88 & 91.80 & 58.90 & 50.42 & 73.18 \\
& NPO+RT ($\mathcal{F}_1$) & 70.83 & 62.82 & 81.47 & 75.12 & 75.03 & 68.77 & 94.64 & 89.26 & 55.20 & 52.73 & 72.84 \\
\cmidrule(lr){2-13}
& GD ($\mathcal{F}_2$) & 82.51 & 54.87 & 58.95 & 44.10 & 87.42 & 68.32 & 93.52 & 80.32 & 56.03 & 50.17 & 74.04 \\
& NPO+RT ($\mathcal{F}_2$) & 93.73 & 85.64 & 89.12 & 83.07 & 93.35 & 87.21 & 95.01 & 88.97 & 58.43 & 52.30 & 75.28 \\
\midrule
\multirow{5}{*}{\rotatebox[origin=c]{90}{LLAMA}}
& Base Model & 88.63 & 84.61 & 88.12 & 76.15 & 90.43 & 83.71 & 96.53 & 92.69 & 64.75 & 75.58 & 55.89 \\
\cmidrule(lr){2-13}
& GD ($\mathcal{F}_1$) & 72.34 & 67.94 & 85.51 & 75.38 & 90.86 & 84.95 & 97.06 & 94.48 & 64.71 & 75.50 & 53.66 \\
& NPO+RT ($\mathcal{F}_1$) & 36.96 & 33.07 & 35.69 & 12.82 & 81.23 & 66.17 & 94.90 & 80.47 & 51.05 & 74.90 & 50.50 \\
\cmidrule(lr){2-13}
& GD ($\mathcal{F}_2$) & 88.11 & 80.25 & 68.27 & 59.23 & 92.45 & 86.99 & 97.89 & 96.57 & 63.38 & 75.37 & 53.92 \\
& NPO+RT ($\mathcal{F}_2$) & 48.25 & 44.35 & 24.43 & 1.79 & 92.02 & 80.76 & 98.78 & 96.27 & 56.42 & 74.67 & 49.23 \\
\bottomrule
\end{tabular}
}
\caption{Results of LoRA unlearning using \textbf{$\mathcal{F}_1$} and \textbf{$\mathcal{F}_2$} for GD and NPO+RT on $\mathcal{D}_{\text{retracted}}$. $\downarrow$ indicates lower is better (forgetting), while $\uparrow$ indicates higher is better (retention and generalization). OLMo-3-7B-Instruct is referred as OLMO and LLAMA3-8B-Instruct is referred as LLAMA.}
\label{tab:main_results_merged_retracted}
\end{table*}

\section{Results}
\label{sec:results}
Table~\ref{tab:main_results_merged} presents the main LoRA results for GD and NPO+RT across both models on $\mathcal{D}_{\text{cs}}$, while additional LoRA methods are reported in App.~\ref{sec:additional_lora_results}. Table~\ref{tab:full_results} in App.~\ref{sec:full_unlearning} presents full-parameter unlearning results for the OLMO model on the dataset $\mathcal{D}_{\text{cs}}$. The full parameter unlearning results for both OLMO and LLAMA models on dataset $\mathcal{D}_{\text{med}}$ and $\mathcal{D}_{\text{retracted}}$ are provided in Table~\ref{tab:main_results_merged_med} and Table~\ref{tab:main_results_merged_retracted} respectively. The results show that forgetting is largely localized to the unlearned forget set, with minimal transfer to the other set. When unlearning is applied to $\mathcal{F}_1$ or $\mathcal{F}_2$, all representative methods significantly reduce performance on the targeted set leaving the counterpart largely unaffected. A complementary evaluation on open-ended questions provides preliminary evidence that the unlearning effect also extends beyond structured QA formats (App.~\ref{sec:free_form_generation}). Qualitative analysis (App.~\ref{sec:qualitative}) indicates that simpler QA formats (e.g., T/F and AR) and higher numbers of paraphrased QAs, and non deeply rooted papers amplify forgetting. However, none of the methods causes sharp performance drops, likely due to LoRA-based fine-tuning \cite{NEURIPS2024_b1f78dfc} and the rootedness of claims in the pretraining corpus. Overall, forgetting remains subset-specific, suggesting current methods suppress surface-level patterns rather than underlying claim-level knowledge. Retention differs across methods: GD, NPO+RT, and SimNPO+RT preserve strong performance on both external and internal retain sets, while NPO and SimNPO show larger retention drops due to the absence of an explicit retain objective. Even in case of LLAMA, there is a significant performance drop in all the sets. This indicates that retain-aware methods better protect non-target knowledge. General benchmarks are affected unevenly: NPO and SimNPO without retention cause notable MMLU degradation in OLMO and HellaSwag degradation in LLAMA, while GD, NPO+RT, and SimNPO+RT largely preserve performance. Reference-model scores show lower drift for semantic than direct members in all $\mathcal{D}_{\text{cs}}$, $\mathcal{D}_{\text{med}}$ and $\mathcal{D}_{\text{retracted}}$, while in case of MIA, Min-K\% AUCs also weaken for semantic members, as shown in detail in Table~\ref{tab:olmo_mia_results}. Details for are provided in App.~\ref{sec:mia}.

\begin{table}[t]
\centering
\caption{Membership inference attack (MIA) results for OLMO using Gradient Descent (GD) unlearning. Higher Ref Score Mean indicates stronger forgetting, while an AUC closer to 0.5 indicates membership indistinguishability (Good Unlearning). Mem indicates Member Data (Target Forget Set) and Sem means Semantic Member Data (Complementary Forget Set)}
\label{tab:olmo_mia_results}
\resizebox{\linewidth}{!}{
\begin{tabular}{llcccc}
\toprule
\multirow{2}{*}{Method} & \multirow{2}{*}{Data} &
\multicolumn{2}{c}{Min-K AUC} &
\multicolumn{2}{c}{Ref Score Mean} \\
\cmidrule(lr){3-4} \cmidrule(lr){5-6}
& & Mem & Sem & Mem & Sem \\
\midrule
GD & CS F1 & 0.68 & 0.73 & 2.89 & 0.57 \\
GD & CS F2  & 0.53 & 0.77 & 3.17 & 0.29 \\
GD & Med F1 & 0.52 & 0.70 & 7.84 & 0.36 \\
GD & Med F2 & 0.39 & 0.77 & 6.61 & 0.93 \\
GD & Retracted F1 & 0.51 & 0.82 & 3.53 & 0.15 \\
GD & Retracted F2 & 0.57 & 0.69 & 4.68 & 0.33 \\
\bottomrule
\end{tabular}
}
\end{table}


\section{Conclusion}

This work introduces Scientific Claim Unlearning and the SciUnlearn benchmark for evaluating whether LMs can forget scientific claims at the claim level, also proving it's applicability with real world retracted claim subset. Experiments show that existing unlearning methods mainly suppress the specific training instances used for unlearning, with limited transfer to paraphrased or complementary forget sets.  Overall, our findings suggest that scientific claim unlearning requires algorithms and evaluations that go beyond surface-form suppression and target structured conceptual knowledge.

\section*{Limitations}

This work has three main limitations. First, we evaluate representative optimization-based unlearning methods, but do not study mechanistic or representation-level unlearning approaches that may better target internal claim representations. Second, although we include a real-world retracted-paper subset, its sample size remains small, limiting the breadth of conclusions that can be drawn from this setting. Expanding this subset with more retracted or explicitly falsified findings across broader scientific domains such as medicine, biology, and chemistry is an important direction for future work. Third, our experiments are limited to 7B--8B models due to the computational cost of full-parameter unlearning, leaving the behavior of larger frontier-scale models for future investigation.

\bibliography{custom}

\appendix

\section{Example of Retracted and Falsified Claims}
\label{sec:motivation_diagram}

\begin{figure*}[t]
    \centering
    \includegraphics[width=1.0\textwidth]{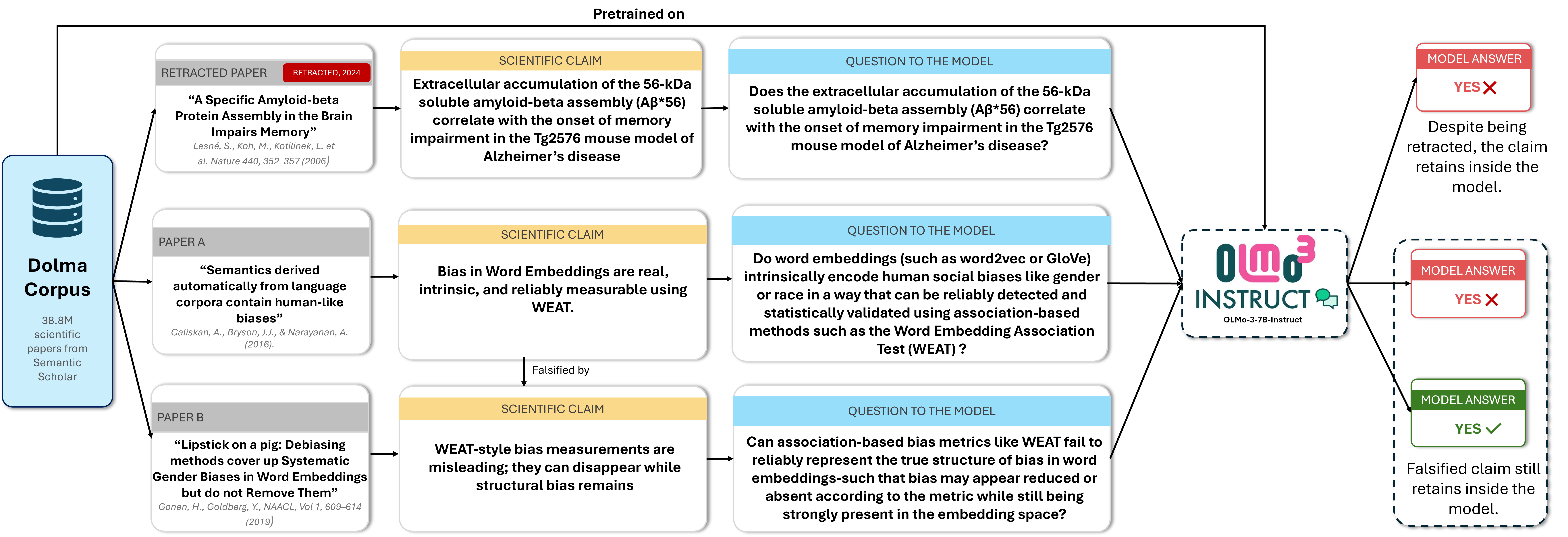}
    \caption{Failure case demonstrating the persistence of invalidated scientific knowledge in language models (OLMO-3-7B-Instruct \cite{groeneveld-etal-2024-olmo}).}
    \label{fig:motivation}
\end{figure*}

\begin{figure*}[ht]
    \centering
    \begin{subfigure}[t]{0.48\textwidth}
        \centering
        \includegraphics[width=\textwidth]{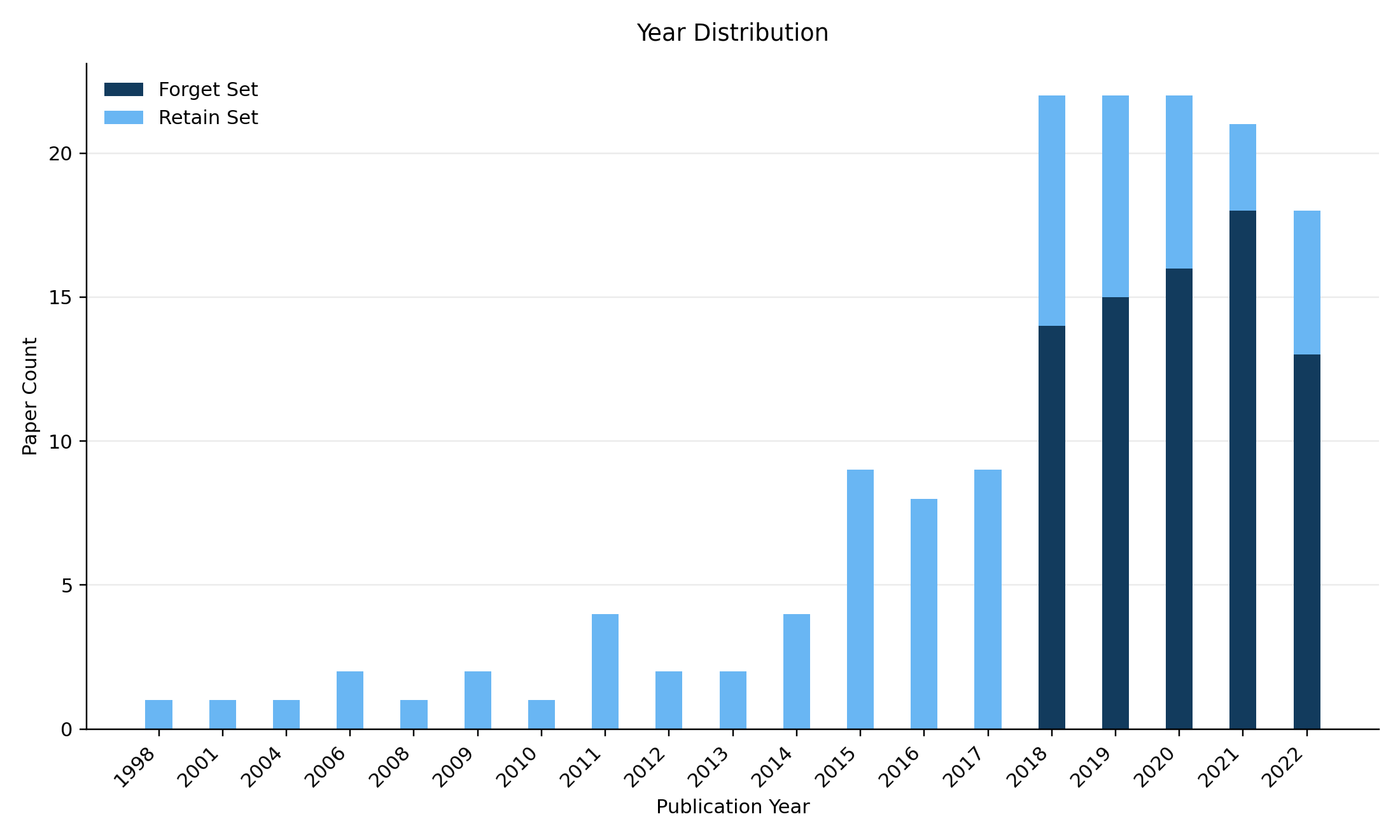}
        \caption{Year-wise distribution of questions}
        \label{fig:compiled_stat_a_cs}
    \end{subfigure}
    \hfill
    \begin{subfigure}[t]{0.48\textwidth}
        \centering
        \includegraphics[width=\textwidth]{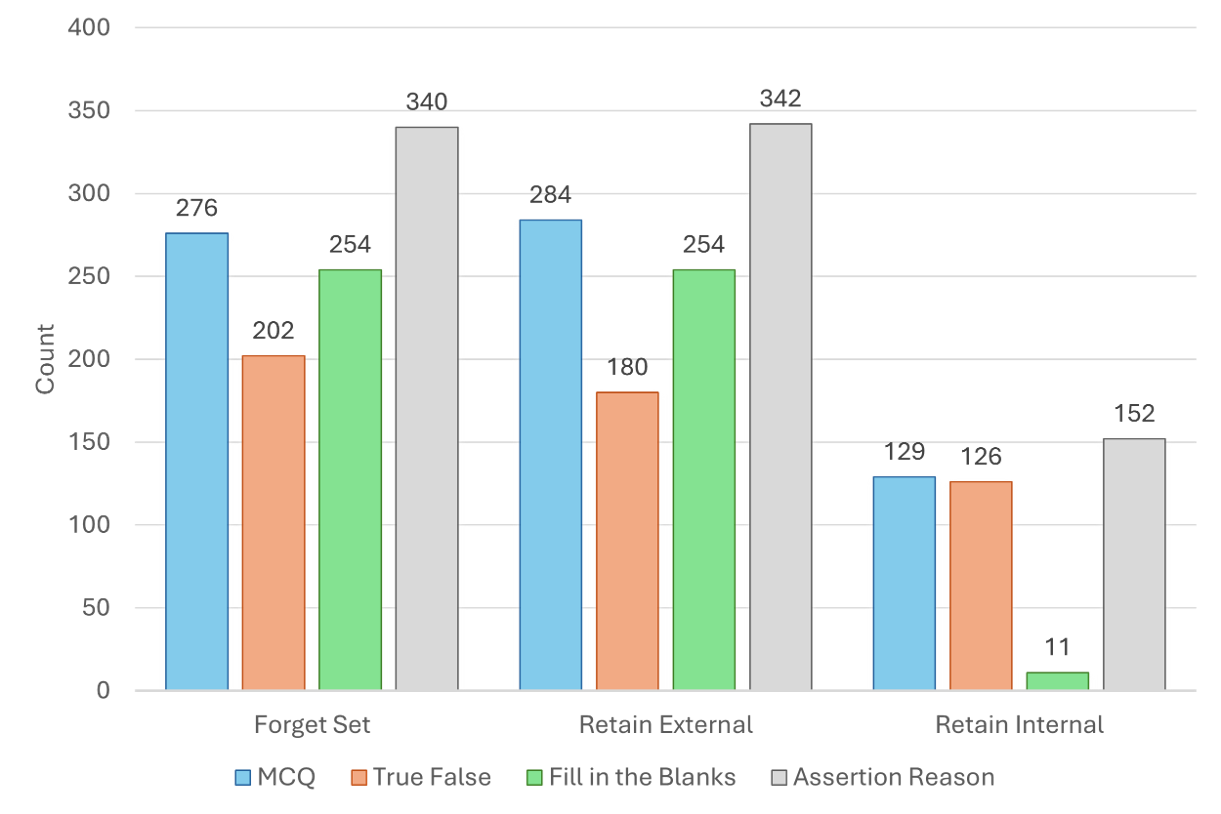}
        \caption{Question type distribution}
        \label{fig:compiled_stat_b_cs}
    \end{subfigure}
    \caption{Dataset statistics for Computer Science subset. (a) Year-wise distribution of questions. (b) Distribution across question types.}
    \label{fig:compiled_stat_cs}
\end{figure*}

\begin{figure*}[ht]
    \centering
    \begin{subfigure}[t]{0.48\textwidth}
        \centering
        \includegraphics[width=\textwidth]{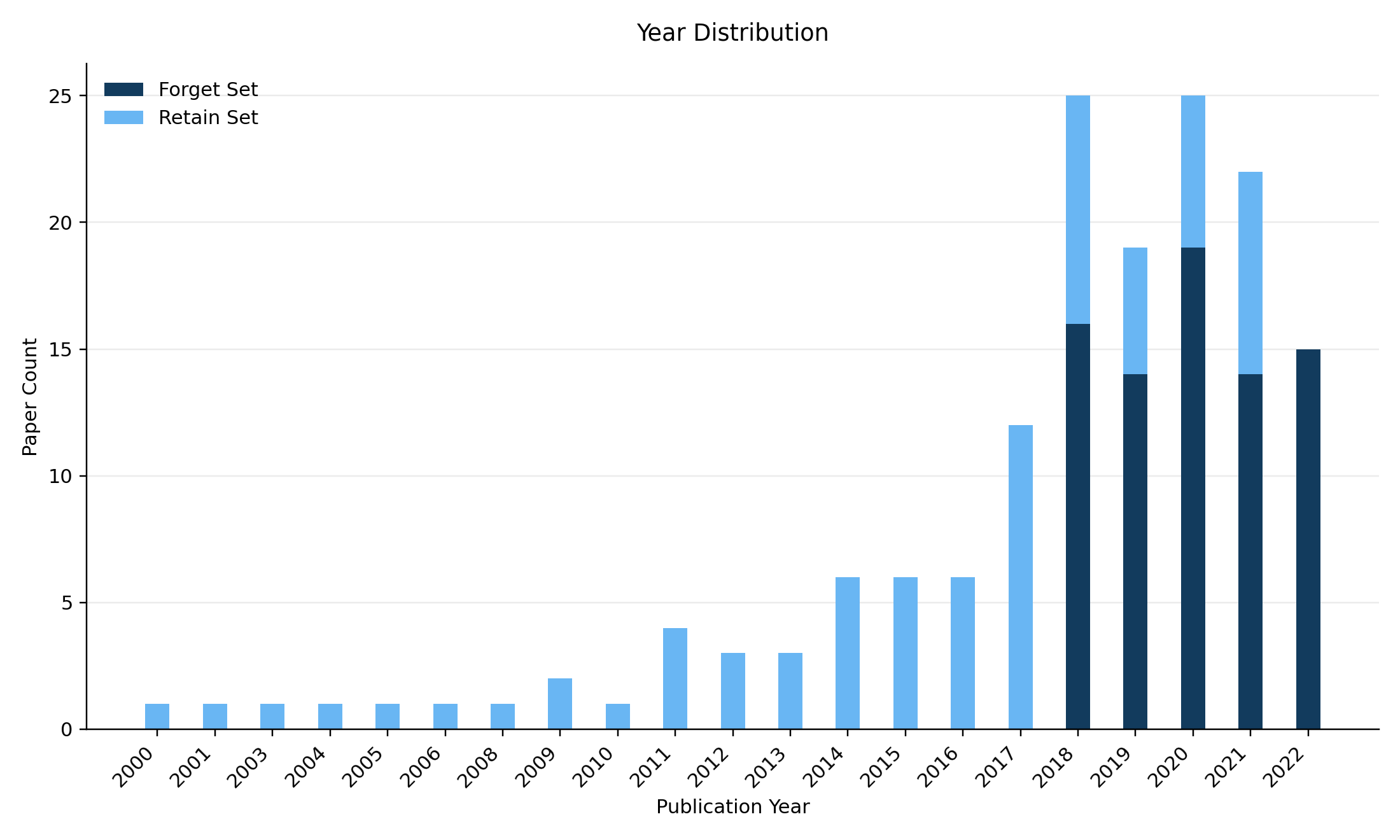}
        \caption{Year-wise distribution of questions}
        \label{fig:compiled_stat_a_med}
    \end{subfigure}
    \hfill
    \begin{subfigure}[t]{0.48\textwidth}
        \centering
        \includegraphics[width=\textwidth]{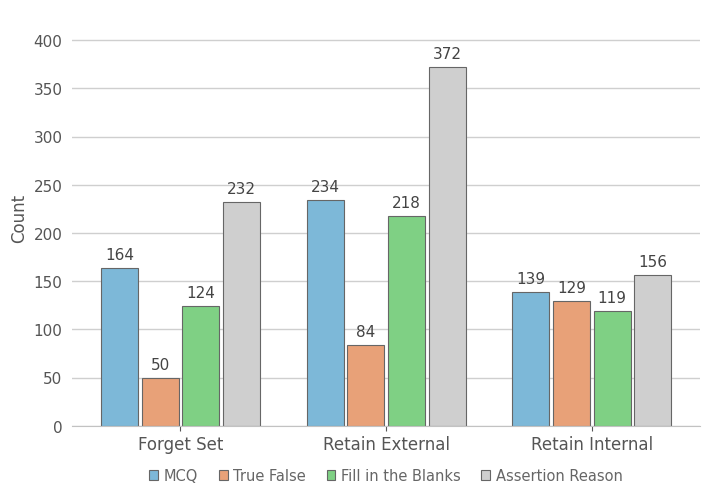}
        \caption{Question type distribution}
        \label{fig:compiled_stat_b_med}
    \end{subfigure}
    \caption{Dataset statistics for Medical subset. (a) Year-wise distribution of questions. (b) Distribution across question types.}
    \label{fig:compiled_stat_med}
\end{figure*}

\begin{figure*}[ht]
    \centering
    \begin{subfigure}[t]{0.48\textwidth}
        \centering
        \includegraphics[width=\textwidth]{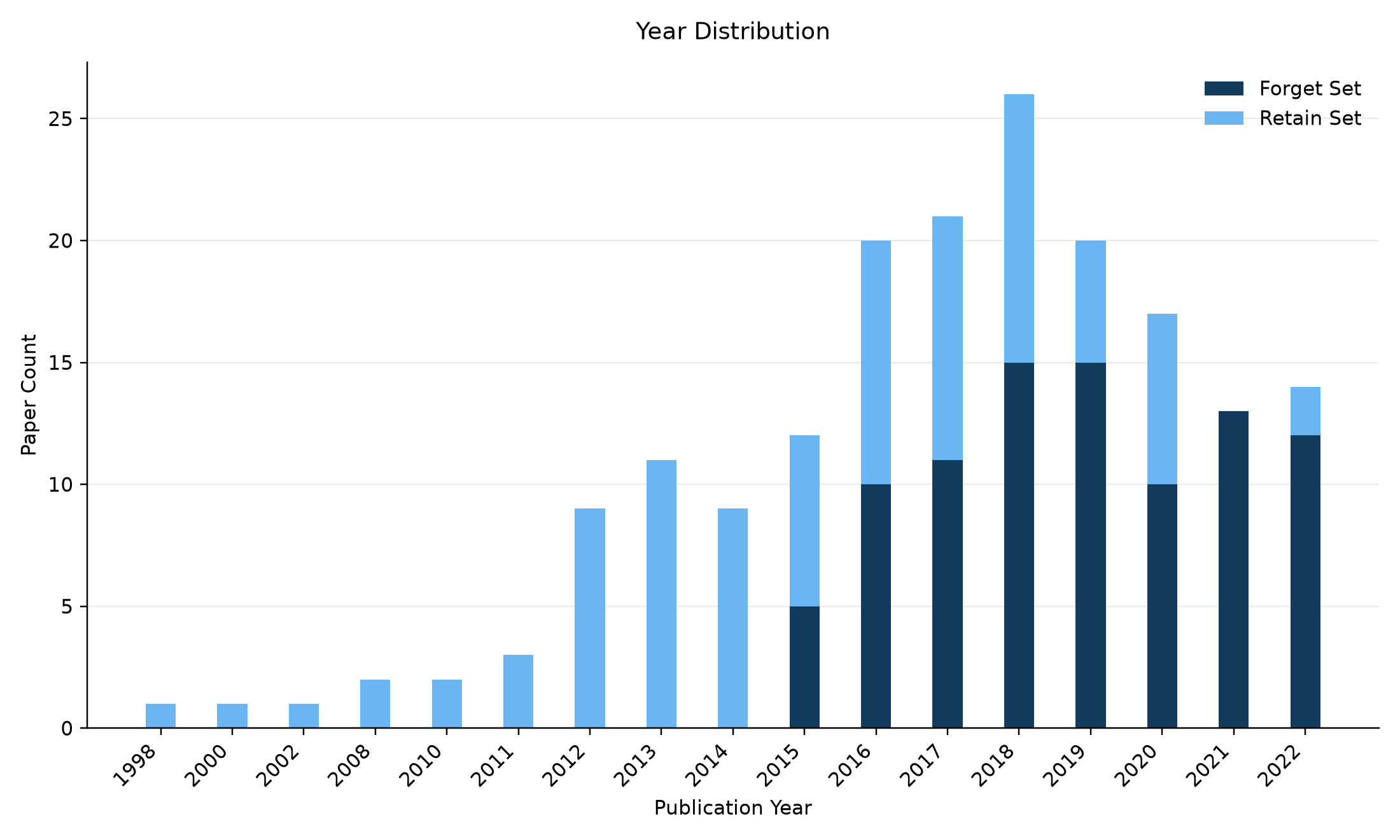}
        \caption{Year-wise distribution of questions}
        \label{fig:compiled_stat_a_ret}
    \end{subfigure}
    \hfill
    \begin{subfigure}[t]{0.48\textwidth}
        \centering
        \includegraphics[width=\textwidth]{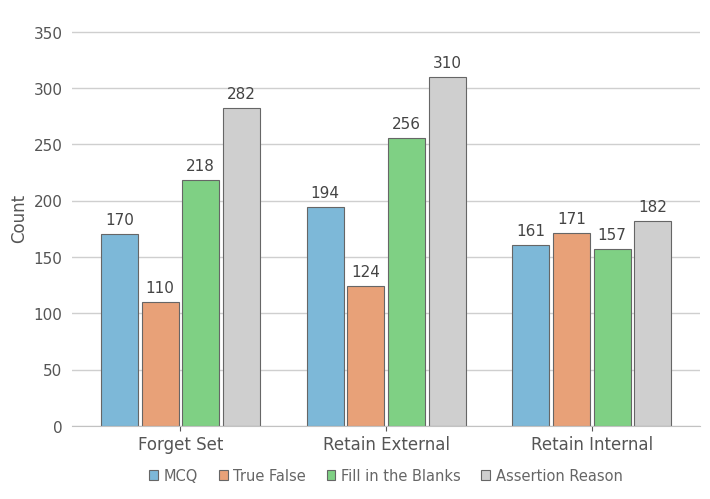}
        \caption{Question type distribution}
        \label{fig:compiled_stat_b_ret}
    \end{subfigure}
    \caption{Dataset statistics for Retracted paper subset. (a) Year-wise distribution of questions. (b) Distribution across question types.}
    \label{fig:compiled_stat}
\end{figure*}

This example illustrates a critical failure mode in modern language models: the persistence of outdated, retracted, or falsified scientific claims. Despite subsequent corrections in the scientific record, models such as OLMO-3-7B-Instruct \cite{groeneveld-etal-2024-olmo} may continue to generate responses that reflect superseded knowledge. As shown in Figure~\ref{fig:motivation}, the model produces confident outputs that do not adequately account for post-publication revisions or retractions.




\section{Human LLM evaluation Scores}
\label{sec:claim_samples}

We present representative examples of scientific claims extracted from papers in the shared github itself in a pdf named \emph{Human LLM Agreement} along with their corresponding LLM-based and human evaluation scores.

\section{Dataset Statistics}
\label{sec:dataset_statistics}

Figure~\ref{fig:compiled_stat_a_cs},\ref{fig:compiled_stat_a_med}, \ref{fig:compiled_stat_a_ret} shows the year‑wise distribution of questions, which is skewed towards more recent papers for both the forget and retain sets. Figure~\ref{fig:compiled_stat_b_cs}, \ref{fig:compiled_stat_b_med}, \ref{fig:compiled_stat_b_ret}  presents the question‑type distribution, illustrating the diversity and relative frequency of different question categories present in the forget and retain sets. The total cost of dataset creation is approximately \$36.57 over 2534 GPT-5 API calls.

\begin{table*}[t]
\centering
\small
\begin{tabularx}{\textwidth}{p{0.18\textwidth} X}
\toprule
\textbf{Field} & \textbf{Content} \\
\midrule
Forget Paper Title & Is it Required? Ranking the Skills Required for a Job-Title \\
\midrule
Extracted Claim & Fine-tuning a multilingual sentence encoder on English job-title data improves skill importance ranking in English while reducing performance on non-English titles. \\
\midrule
Verbatim Claim & The model also performs fairly on other languages (as seen in the table 3), but by fine tuning we lose the multilingual capabilities of the ranking system. \\
\bottomrule
\end{tabularx}
\caption{Forget-set claim used for sample QA generation from Computer Science subset.}
\label{tab:sample_forget_claim}
\end{table*}

\begin{table*}[t]
\centering
\tiny
\setlength{\tabcolsep}{3pt}
\renewcommand{\arraystretch}{0.95}
\begin{tabularx}{\textwidth}{p{0.12\textwidth} X p{0.08\textwidth} p{0.08\textwidth} p{0.14\textwidth}}
\toprule
\textbf{Type} & \textbf{Question} & \textbf{Reference} & \textbf{OLMo} & \textbf{Metrics} \\
\midrule
MCQ 1 & Choose the correct option (A--D): When a multilingual sentence encoder is fine-tuned on English job-title data, what is the observed effect on skill importance ranking for English and non-English titles? [A] Both English and non-English performance improve equally [B] English skill importance ranking improves, but performance on non-English titles decreases [C] English performance decreases while non-English improves [D] There is no measurable change in either English or non-English performance & B & B & EM=1.0, ROUGE=1.0, SemSim=1.0 \\
\midrule
MCQ 2 & Choose the correct option (A--D): Fine-tuning a multilingual sentence encoder using only English job-title data leads to which outcome in skill importance ranking across languages? [A] Improvement limited to non-English titles with no change in English [B] Degradation for both English and non-English titles [C] Improved English results coupled with reduced performance on non-English titles [D] No trade-off; both English and non-English results improve & C & C & EM=1.0, ROUGE=1.0, SemSim=1.0 \\
\midrule
True/False 1 & Fine-tuning a multilingual sentence encoder on English job-title data increases skill importance ranking performance for English titles while lowering performance for non-English titles. & True & True & EM=1.0, ROUGE=1.0, SemSim=1.0 \\
\midrule
True/False 2 & Fine-tuning a multilingual sentence encoder on English job-title data improves skill importance ranking performance for both English and non-English titles. & False & False & EM=1.0, ROUGE=1.0, SemSim=1.0 \\
\midrule
Fill-in-the-blank 1 & When a multilingual sentence encoder is fine-tuned on English job-title data, English skill importance ranking improves, but \_\_\_\_\_\_\_\_\_ (non-English performance decreases/non-English performance also improves). & non-English performance decreases & non-English performance decreases & EM=1.0, ROUGE=1.0, SemSim=1.0 \\
\midrule
Fill-in-the-blank 2 & Fine-tuning a multilingual sentence encoder using English job-title data produces a trade-off: better English skill importance ranking and \_\_\_\_\_\_\_\_\_ (worse non-English performance/unchanged non-English performance). & worse non-English performance & worse non-English performance & EM=1.0, ROUGE=1.0, SemSim=1.0 \\
\midrule
Assertion--Reason 1 & Assertion (A): Fine-tuning a multilingual sentence encoder on English job-title data improves skill importance ranking for English titles while reducing performance on non-English titles. Reason (R): The fine-tuning uses only English job-title examples, not multilingual data. & A is True, R is True, and R explains A. & A is True, R is True, and R explains A. & EM=1.0, ROUGE=1.0, SemSim=1.0 \\
\midrule
Assertion--Reason 2 & Assertion (A): Fine-tuning a multilingual sentence encoder on English job-title data yields improved English skill importance ranking alongside reduced performance on non-English titles. Reason (R): Because the model is perfectly multilingual, English-only fine-tuning cannot hurt non-English performance. & A is True, R is False & A is True, R is False & EM=1.0, ROUGE=1.0, SemSim=1.0 \\
\bottomrule
\end{tabularx}
\renewcommand{\arraystretch}{1.0}
\caption{Sample forget-set QA pairs generated from the claim in Table~\ref{tab:sample_forget_claim}.}
\label{tab:sample_forget_qa}
\end{table*}

\begin{table*}[t]
\centering
\small
\begin{tabularx}{\textwidth}{p{0.18\textwidth} X}
\toprule
\textbf{Field} & \textbf{Content} \\
\midrule
Retain Paper Title & JobBERT: Understanding Job Titles through Skills \\
\midrule
Extracted Claim & Semantic representations of job titles learned by a BERT-based encoder trained with distant supervision to predict co-occurring skills from vacancy texts, using token-level gating and negative sampling, enable taxonomy-agnostic nearest-neighbor normalization without manual title labels and outperform generic sentence encoders on this task. \\
\bottomrule
\end{tabularx}
\caption{Retain-set claim used for sample QA generation from Computer Science subset.}
\label{tab:sample_retain_claim}
\end{table*}

\begin{table*}[t]
\centering
\tiny
\setlength{\tabcolsep}{3pt}
\renewcommand{\arraystretch}{0.95}
\begin{tabularx}{\textwidth}{p{0.12\textwidth} X p{0.08\textwidth} p{0.08\textwidth} p{0.14\textwidth}}
\toprule
\textbf{Type} & \textbf{Question} & \textbf{Reference} & \textbf{OLMo} & \textbf{Metrics} \\
\midrule
MCQ 1 & Choose the correct option (A--D): Which statement correctly describes how semantic representations of job titles are learned and what they achieve? [A] Representations are learned by a BERT-based encoder trained with distant supervision to predict co-occurring skills from vacancy texts, using token-level gating and negative sampling; they enable taxonomy-agnostic nearest-neighbor normalization without manual title labels and outperform generic sentence encoders on this task. [B] Representations are learned by an LSTM-based encoder trained with manual title labels, without token-level gating or negative sampling; they enable taxonomy-specific normalization via rule-based matching and do not outperform generic sentence encoders on this task. [C] Representations are learned by a BERT-based encoder trained to predict job titles from skills, using token-level gating but no negative sampling; they require manual title labels for normalization and show performance comparable to generic sentence encoders. [D] Representations are obtained by unsupervised averaging of word embeddings from vacancy texts; they support taxonomy-agnostic cluster-based normalization but only match the performance of generic sentence encoders on this task. & A & A & EM=1.0, ROUGE=1.0, SemSim=1.0 \\
\midrule
MCQ 2 & Choose the correct option (A--D): What configuration yields job-title representations that enable taxonomy-agnostic nearest-neighbor normalization without manual title labels and surpass generic sentence encoders on that task? [A] Unsupervised averaging of token embeddings from vacancy texts with cluster-based normalization and no negative sampling. [B] A BERT-based encoder trained to reconstruct titles from taxonomies using manual labels, with attention but no negative sampling. [C] A BERT-based encoder trained with distant supervision to predict co-occurring skills from vacancy texts, using token-level gating and negative sampling. [D] A generic sentence encoder fine-tuned on unrelated corpora with taxonomy-specific nearest-centroid mapping. & C & C & EM=1.0, ROUGE=1.0, SemSim=1.0 \\
\midrule
True/False 1 & Semantic representations of job titles learned by a BERT-based encoder trained with distant supervision to predict co-occurring skills from vacancy texts, using token-level gating and negative sampling, enable taxonomy-agnostic nearest-neighbor normalization without manual title labels and outperform generic sentence encoders on this task. & True & True & EM=1.0, ROUGE=1.0, SemSim=1.0 \\
\midrule
True/False 2 & Semantic representations of job titles learned by a BERT-based encoder trained with distant supervision to predict co-occurring skills from vacancy texts, without token-level gating or negative sampling, require manual title labels for taxonomy-specific nearest-neighbor normalization and do not outperform generic sentence encoders on this task. & False & False & EM=1.0, ROUGE=1.0, SemSim=1.0 \\
\midrule
Fill-in-the-blank 1 & The learned job-title representations support \_\_\_\_\_\_\_\_\_ (taxonomy-agnostic/taxonomy-specific) nearest-neighbor normalization without manual title labels and outperform generic sentence encoders, given they are produced by a BERT-based encoder trained with distant supervision to predict co-occurring skills from vacancy texts using token-level gating and negative sampling. & taxonomy-agnostic & taxonomy-agnostic & EM=1.0, ROUGE=1.0, SemSim=1.0 \\
\midrule
Fill-in-the-blank 2 & Semantic representations of job titles produced by a \_\_\_\_\_\_\_\_\_ (BERT-based/LSTM-based) encoder trained with distant supervision to predict co-occurring skills from vacancy texts, with token-level gating and negative sampling, enable taxonomy-agnostic nearest-neighbor normalization without manual title labels and outperform generic sentence encoders on this task. & worse non-English performance & worse non-English performance & EM=1.0, ROUGE=1.0, SemSim=1.0 \\
\midrule
Assertion--Reason 1 & Assertion (A): Semantic representations of job titles learned by a BERT-based encoder trained with distant supervision to predict co-occurring skills from vacancy texts, using token-level gating and negative sampling, enable taxonomy-agnostic nearest-neighbor normalization without manual title labels and outperform generic sentence encoders on this task. Reason (R): Predicting co-occurring skills from vacancy texts under distant supervision aligns representations with job-title semantics, and token-level gating plus negative sampling increases discriminative power. & A is True, R is True, and R explains A. & A is True, R is True, and R explains A. & EM=1.0, ROUGE=1.0, SemSim=1.0 \\
\midrule
Assertion--Reason 2 & Assertion (A): Semantic representations of job titles learned by a BERT-based encoder trained with distant supervision to predict co-occurring skills from vacancy texts, using token-level gating and negative sampling, enable taxonomy-agnostic nearest-neighbor normalization without manual title labels and outperform generic sentence encoders on this task. Reason (R): Manual title labels and taxonomy-specific rules are required for training and normalization in this setting. & A is True, R is False & A is True, R is False & EM=1.0, ROUGE=1.0, SemSim=1.0 \\
\bottomrule
\end{tabularx}
\renewcommand{\arraystretch}{1.0}
\caption{Sample retain-set QA pairs generated from the claim in Table~\ref{tab:sample_retain_claim}.}
\label{tab:sample_retain_qa}
\end{table*}

\section{Sample Question-Answer}
\label{sec:sample_qa_eval}

This section presents qualitative examples illustrating the end-to-end pipeline, including extracted scientific claims (Table~\ref{tab:sample_forget_claim} and \ref{tab:sample_retain_claim}), generated question-answer pairs (Table~\ref{tab:sample_forget_qa} and \ref{tab:sample_retain_qa}), and their evaluation from Computer Science subset. For each paper, we list the extracted claims, followed by representative question-answer pairs derived from a selected claim, verbatim statement from where the claim is taken, along with the reference answers, model predictions, and evaluation metrics for both forget and retain set.

\section{Distribution of Semantic Overlap Strength}
\label{sec:overlap}

\begin{figure}[ht]
\centering
\includegraphics[width=0.90\columnwidth]{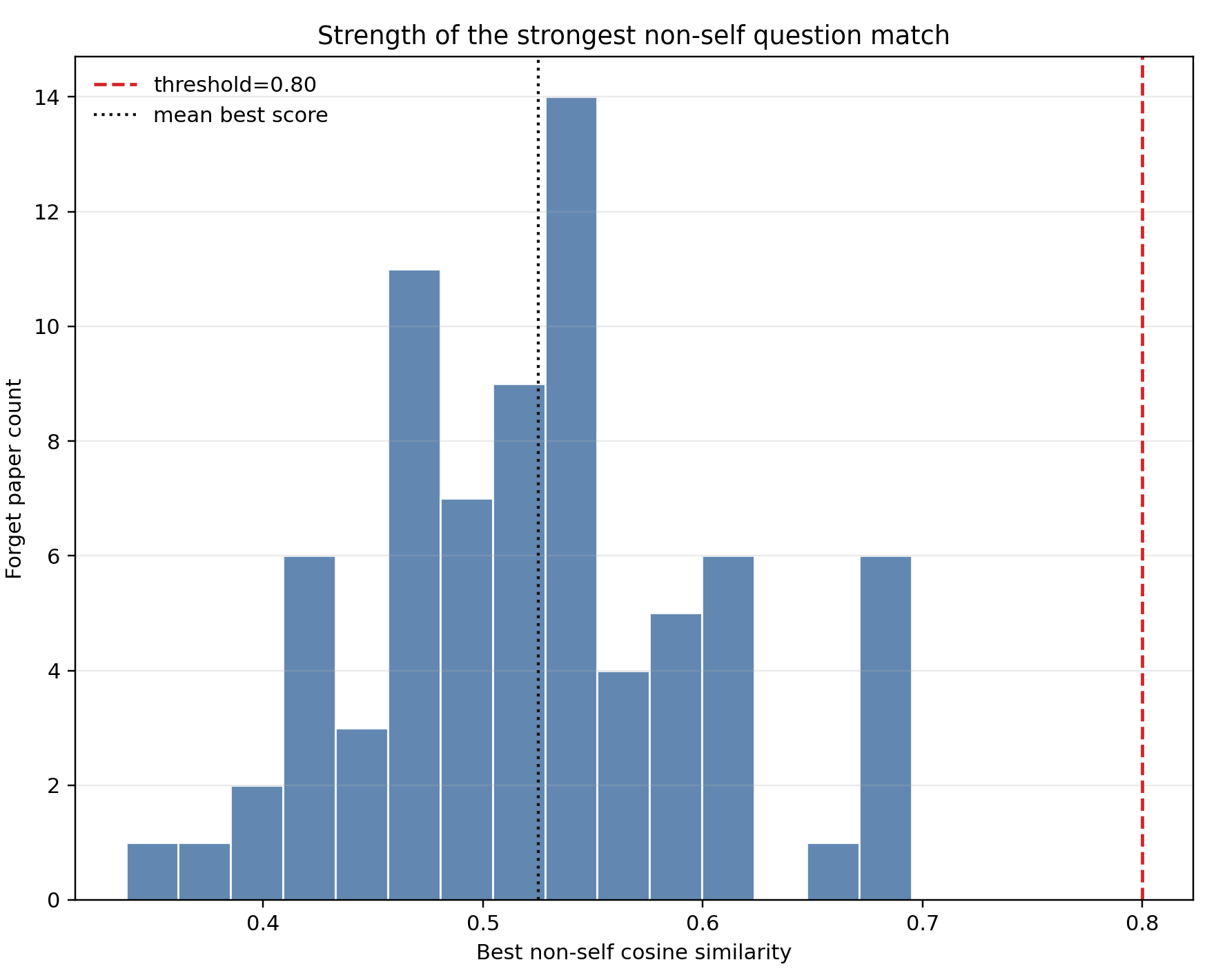}
\caption{Distribution of the strongest non-self cosine similarity for each forget paper. The red dashed line indicates the overlap threshold ($\tau = 0.8$), and the black dotted line denotes the mean similarity.}
\label{fig:overlap_hist}
\end{figure}

We analyze the extent of semantic overlap between the forget and retain sets, the distribution of the strongest non-self question similarity for each forget paper is examined. Here, non‑self denotes all retain papers excluding the retain paper that directly corresponds to the anchor paper itself. For each forget paper, the maximum cosine similarity between any of its questions and all questions from non-corresponding retain papers was computed. This yields a single ``best non-self similarity'' score per paper, capturing the strongest potential semantic overlap. Figure~\ref{fig:overlap_hist} shows the distribution of these scores across all forget papers. An empirical threshold of $\tau = 0.8$ is used to define significant semantic overlap.

The distribution is concentrated in the range of approximately $0.45$ to $0.65$, with a mean similarity around $0.53$. Importantly, no forget paper reaches or exceeds the threshold of $0.8$, indicating the absence of high-confidence semantic overlap between the forget and retain sets. The absence of high-similarity matches confirms that the retain set is effectively disjoint from the forget set at a semantic level. This validates the dataset construction process and ensures that any observed unlearning behavior cannot be attributed to direct leakage or duplication across the two sets.

\section{Implementation Details}
\label{sec:implementation}

\subsection{Gradient Difference (GD)}

Gradient Difference modifies model parameters by performing gradient descent on the retain set while simultaneously performing gradient ascent on the forget set, thereby reducing the likelihood of generating target forget data \cite{Neel2020DescenttoDeleteGM}. The update rule is defined in Equation \ref{eq:gradient_difference}.
\begin{equation}
\theta' = \theta - \eta \, w_r \nabla_{\theta} \mathcal{L}_{retain}(\theta) + \eta \, w_f \nabla_{\theta} \mathcal{L}_{forget}(\theta)
\label{eq:gradient_difference}
\end{equation}

where $\theta$ represents model parameters, $\eta$ is the learning rate, $\mathcal{L}_{forget}$ denotes the loss on the forget set, and $\mathcal{L}_{retain}$ denotes the loss on a retain set to preserve useful knowledge.

\subsection{Negative Preference Optimization (NPO)}

Negative Preference Optimization \cite{Zhang2024NegativePO, bronec-helcl-2025-atyaephyra} formulates unlearning as a preference learning problem, where the model is encouraged to prefer retained responses over those associated with the forget set. The objective is given in Equation \ref{eq:npo}.
\begin{equation}
\mathcal{L}_{\text{NPO}} = - \frac{2}{\beta} \, \mathbb{E}_{(x,y_f)} \left[ 
\log \sigma \left( \beta \, \log \frac{p_{\theta_0}(y_f|x)}{p_{\theta}(y_f|x)} \right)
\right]
\label{eq:npo}
\end{equation}

where $p_{\theta}(y|x)$ denotes the conditional likelihood of generating response $y$ given input $x$ under the model parameterized by $\theta$, and $p_{\theta_0}(y|x)$ denotes the corresponding likelihood under the reference (pretrained) model, which remains fixed during training. $y_f$ denotes responses corresponding to the forget set, $\sigma(\cdot)$ is the sigmoid function, $\beta$ is a scaling parameter controlling the sharpness of the preference.

\subsection{NPO with Retain Set Fine-Tuning}

To improve stability and mitigate unintended degradation of useful knowledge, the NPO objective is augmented with an additional retain set optimization. While NPO suppresses the likelihood of forget samples relative to a reference model, it does not explicitly enforce the preservation of retained knowledge. Therefore, a complementary retain objective is introduced.

The retain loss is defined in Equation \ref{eq:retain_npo}.
\begin{equation}
\mathcal{L}_{retain} = - \mathbb{E}_{(x,y_r)} \left[ \log p_{\theta}(y_r|x) \right]
\label{eq:retain_npo}
\end{equation}

The overall training objective is given in Equation \ref{eq:npo_rt}.
\begin{equation}
\mathcal{L} = \lambda_{\text{NPO}} \mathcal{L}_{\text{NPO}} + \lambda_{r} \mathcal{L}_{retain}
\label{eq:npo_rt}
\end{equation}

where $\lambda_{\text{NPO}}$ and $\lambda_{r}$ control the relative contributions of the unlearning and retention objectives, respectively.

\begin{table*}[ht]
\centering
\small
\setlength{\tabcolsep}{3pt}
\begin{tabular}{lcc|cc|cc|cc|cc}
\hline
\textbf{Param} 
& \multicolumn{2}{c|}{\textbf{GD}} 
& \multicolumn{2}{c|}{\textbf{NPO}} 
& \multicolumn{2}{c|}{\textbf{NPO+RT}} 
& \multicolumn{2}{c|}{\textbf{SimNPO}} 
& \multicolumn{2}{c}{\textbf{SimNPO+RT}} \\

& $\mathcal{F}_1$ & $\mathcal{F}_2$
& $\mathcal{F}_1$ & $\mathcal{F}_2$
& $\mathcal{F}_1$ & $\mathcal{F}_2$
& $\mathcal{F}_1$ & $\mathcal{F}_2$
& $\mathcal{F}_1$ & $\mathcal{F}_2$ \\
\hline

LR 
& $2e-5$ & $1e-5$
& $1e-5$ & $1e-5$
& $1e-5$ & $1e-5$
& $2e-5$ & $2e-5$
& $2e-5$ & $2e-5$ \\

Epochs 
& 50 & 10
& 10 & 10
& 30 & 15
& 15 & 15
& 30 & 10 \\

$\beta$
& -- & --
& 0.3 & 0.3
& 0.3 & 0.3
& 0.3 & 0.3
& 0.3 & 0.3 \\

$\gamma$
& -- & --
& -- & --
& -- & --
& 0.0 & 0.0
& 0.0 & 0.0 \\

$w_f, w_r$
& (1.0, 3.0) & (1.2, 1.5)
& (1.0, 0.0) & (1.0, 0.0)
& (1.0, 4.0) & (1.2, 0.75)
& (1.0, 4.0) & (1.0, 4.0)
& (1.0, 4.0) & (1.0, 0.25)\\

\hline
\end{tabular}
\caption{Hyperparameters for unlearning methods under forget sets $\mathcal{F}_1$ and $\mathcal{F}_2$. LR is Learning Rate;
$\beta$ is preference strength in NPO;  
$\gamma$ scales retain regularization in SimNPOs;  
$w_f$ and $w_r$ is weights for forget and retain objectives.}
\label{tab:hyperparams}
\end{table*}

\subsection{SimNPO (Simplified Negative Preference Optimization)}

SimNPO \cite{Fan2024SimplicityPR} is a simplified variant of preference-based unlearning that operates without a reference model or explicit retain objective. Unlike standard NPO, which compares the current model against a fixed reference, SimNPO directly penalizes the likelihood of generating forget samples, making it a fully self-contained and reference-free unlearning approach. The objective is defined in Equation~\ref{eq:simnpo}.

{
\small 
\begin{equation}
\mathcal{L}_{\text{SimNPO}} = - \frac{2}{\beta} \, \mathbb{E}_{(x,y_f)} \left[
\log \sigma \left( - \beta \, \log p_{\theta}(y_f|x) - \gamma \right)
\right],
\label{eq:simnpo}
\end{equation}
}

where $ p_{\theta}(y_f|x)$ denotes the average log-likelihood of the response $y_f$ given input $x$, computed over answer tokens only. Here, $\sigma(\cdot)$ is the sigmoid function, $\beta$ controls the sharpness of the penalty, and $\gamma$ is a margin term.

This formulation directly encourages the model to reduce the likelihood of forget samples without relying on external comparisons or paired preferences. By operating on the average log-probability over answer tokens, SimNPO focuses on suppressing the generation of target responses at a sequence level.

\subsection{SimNPO with Retain Set Fine-Tuning (SimNPO+RT)}

To alleviate unintended forgetting and better preserve useful knowledge, SimNPO is extended with an explicit retain set objective. While SimNPO effectively suppresses forget samples by directly penalizing their likelihood, it does not impose any constraint to maintain performance on retained data. To address this limitation, a retain loss is incorporated alongside the SimNPO objective.

The retain loss is defined identically to Equation~\ref{eq:retain_npo}.Thus the combined SimNPO+RT objective is then given in Equation~\ref{eq:simnpo_rt}
\begin{equation}
\mathcal{L} = \lambda_{\text{SimNPO}} \mathcal{L}_{\text{SimNPO}} + \lambda_{r} \mathcal{L}_{\text{retain}},
\label{eq:simnpo_rt}
\end{equation}

Here, $\lambda_{\text{SimNPO}}$ and $\lambda_{r}$ balance the contributions of the unlearning and retention objectives, respectively. By explicitly encouraging correct predictions on the retain set, SimNPO+RT improves training stability and helps maintain overall model utility while remaining free of reference-model comparisons.

\section{Hyperparameter Selection for Algorithm Implementation}
\label{sec:hyperpaarmeter}

As shown in Table~\ref{tab:hyperparams}, hyperparameters are chosen specifically to balance the trade-off between forgetting and retention, ensuring that the unlearning process effectively removes targeted information while preserving performance on retained data. Values are selected to achieve sufficient forgetting without inducing over-forgetting, which would unnecessarily degrade retained knowledge, thereby maintaining an optimal equilibrium between removal efficacy and utility preservation. Besides, different algorithms require distinct hyperparameters due to inherent differences in their unlearning dynamics and sensitivity to the forgetting–retention trade-off.

\section{Distinguishing Memorization from General Reasoning}
\label{sec:memorization_reasoning}

Correct QA performance alone does not establish that a model has memorized the underlying content, particularly for multiple-choice questions that may sometimes be answered through general reasoning or surface-level cues. We therefore conduct two complementary analyses to assess whether SciUnlearn measures knowledge acquired during pretraining rather than reasoning ability alone.

\paragraph{Likelihood-based membership signals.}
Following prior work on detecting pretraining data \cite{Shi2023DetectingPretraining}, we compare benchmark samples sourced from Dolma (members) with 97 samples drawn from papers published in 2025 and therefore absent from the Dolma corpus (non-members). The two groups exhibit a clear separation. Member samples have a substantially lower average negative log-likelihood (NLL) than non-members (0.24 vs.\ 3.88) and a higher Min-K\% token-probability score at $K=20\%$ (0.73 vs.\ 0.47). Thus, OLMo assigns markedly higher likelihood to the benchmark samples, providing evidence that their content was encountered during pretraining.

\paragraph{Robustness to answer-order perturbations.}
We additionally test whether MCQ performance depends on answer-position cues. For a subset of 10 four-option questions, we enumerate all $4!=24$ permutations of the answer choices, producing 240 variants in total, following the answer-order intervention used by \citet{Muhamed2025SAEs}. Before unlearning, the model answers 227 of the 240 variants correctly (94.6\%), consistently selecting the same answer content despite changes in its position. This result rules out a fixed answer-position preference and demonstrates stable access to question-specific knowledge across nearly all orderings.

After unlearning, accuracy falls to 114 of 240 variants (47.5\%), a decrease of 47.1 percentage points. Because this reduction persists across all answer positions, it is unlikely to be explained solely by a change in positional preference. Instead, the result indicates that unlearning weakens the model's ability to retrieve the targeted information consistently.

Neither analysis alone can definitively separate memorization from every form of generalization. Taken together, however, the strong member--non-member separation in likelihood-based signals and the large, order-invariant decline after unlearning provide independent evidence that SciUnlearn captures knowledge acquired from the pretraining corpus and is therefore an appropriate setting for evaluating scientific claim unlearning.

\section{Free-Form Generation Evaluation}
\label{sec:free_form_generation}

To examine whether the observed unlearning effects extend beyond structured QA formats, we conduct a small-scale free-form generation experiment. We construct 50 open-ended questions from 50 claims in the computer science dataset and apply Gradient Difference (GD) unlearning to OLMo. Because exact-match and ROUGE scores are poorly suited to open-ended responses with multiple valid surface forms, we evaluate the likelihood assigned to the target answers using negative log-likelihood (NLL) and Min-K\% token probability.

\begin{table}[ht]
\centering
\small
\setlength{\tabcolsep}{3pt}
\begin{tabular}{lcc}
\toprule
\textbf{Metric} & \textbf{Base Model} & \textbf{Unlearned Model} \\
\midrule
Mean NLL                    & 1.39 & 2.59 \\
Median NLL                  & 1.37 & 2.19 \\
Mean Min-K\% prob.          & 0.58 & 0.47 \\
Median Min-K\% prob.        & 0.57 & 0.48 \\
\bottomrule
\end{tabular}
\caption{Free-form generation results on 50 open-ended questions before and after GD unlearning. Higher NLL and lower Min-K\% probability indicate reduced likelihood of the target information.}
\label{tab:free_form_generation}
\end{table}

As shown in Table~\ref{tab:free_form_generation}, unlearning increases both mean and median NLL, while decreasing the corresponding Min-K\% token-probability statistics. The mean NLL rises from 1.39 to 2.59 and the mean Min-K\% probability falls from 0.58 to 0.47, indicating that the unlearned model assigns lower likelihood to the targeted information. Although this experiment is limited to 50 questions and one model--algorithm configuration, it provides preliminary evidence that the unlearning effect extends beyond structured QA evaluation to free-form generation.

\section{Reference Score Drift and Membership Inference Attack}
\label{sec:mia}

To further analyze whether unlearning transfers beyond the explicitly targeted forget subset, we conduct an additional membership inference analysis \cite{Duan2024DoMI} using both direct members and \textit{semantic members}. Due to computational and time constraints, we restrict this analysis to the strongest-performing configuration in our experiments, namely GD, NPO+RT and SimNPO+RT on $\mathcal{F}_1$. Unlike conventional MIA settings, our goal is not only to distinguish members from non-members, but also to study whether semantically related claims from the complementary forget subset $\mathcal{F}_2$ exhibit membership-like behavior after unlearning. This allows us to evaluate whether forgetting propagates beyond the directly unlearned samples.

Specifically, we define three groups:
(i) \textbf{Member}: QA pairs originating from the directly unlearned forget subset $\mathcal{F}_1$,
(ii) \textbf{Semantic Member}: semantically related QA pairs from $\mathcal{F}_2$, and
(iii) \textbf{Non-member}: unrelated external QA samples not involved in training or unlearning.

The semantic-member setup differs from traditional MIA formulations, where the objective is usually binary member vs.\ non-member discrimination. Here, semantic members are intentionally introduced to probe the locality and transferability of forgetting. 

We evaluate two signals: Min-K\% probability and reference-model score.

\paragraph{(1) Min-K\% Probability}
We further compute the Min-K\% score proposed in prior memorization studies:
\[
\text{MinK}(x)
=
\frac{1}{|S_k|}
\sum_{t \in S_k}
\log p_\theta(y_t \mid x, y_{<t}),
\]
where $S_k$ denotes the bottom-$k\%$ least confident output tokens. We use $k=20\%$ following prior work. Less negative values indicate stronger membership behavior.

\paragraph{(2) Reference Model Score}
To quantify drift relative to the original pretrained model, we compute:
\[
\text{RefScore}(x)
=
\mathcal{L}_{\text{unlearned}}(x)
-
\mathcal{L}_{\text{base}}(x).
\]
Scores near zero indicate that the unlearned model behaves similarly to the original base model, whereas larger positive values indicate stronger forgetting-induced deviation.

Table~\ref{tab:olmo_mia_results} shows that direct members exhibit stronger forgetting signatures than semantic members. Overall, the MIA results indicate that forgetting is mostly localized to $\mathcal{F}_1$ and transfers only weakly to semantically related claims in $\mathcal{F}_2$ and vice versa.

\section{Examples of Unlearning Across Algorithms}
\label{sec:qa_algo}

To demonstrate the effectiveness of unlearning, we present representative examples from the forget set in Table~\ref{tab:unlearning_example} for Computer Science subset. All algorithms consistently produce incorrect outputs, indicating successful forgetting.

\begin{table*}[ht]
\small

\begin{tabular}{p{\textwidth}} \\
\toprule 
\textbf{Example 1} \\
\midrule
\textbf{Question:} For Z24 Bridge natural frequency data, time-delay embeddings of the raw series exhibit a toroidal topology driven by cyclic temperature effects, whereas embeddings of cointegrated residuals exhibit an open-ball topology characteristic of approximately Gaussian white-noise processes. \\
\textbf{Ground Truth:} True \\
\textbf{GD:} False \quad
\textbf{NPO:} False \quad
\textbf{NPO+RT:} False \quad
\textbf{SimNPO:} False \quad
\textbf{SimNPO+RT:} False \\
\midrule

\textbf{Example 2} \\
\midrule
\textbf{Question:} When the second natural frequency depends nonlinearly on temperature, linear cointegration leaves residual topological structure, whereas Gaussian process-based nonlinear cointegration trained on data that include the nonlinear regime more effectively removes these effects. \\
\textbf{Ground Truth:} True \\
\textbf{GD:} False \quad
\textbf{NPO:} False \quad
\textbf{NPO+RT:} False \quad
\textbf{SimNPO:} False \quad
\textbf{SimNPO+RT:} False \\
\midrule

\textbf{Example 3} \\
\midrule
\textbf{Question:} In a linear classifier with an additive patch trigger, the optimal untargeted adversarial perturbation has a \underline{\hspace{0.8cm}}(substantial projection/near-zero projection) onto the trigger direction, theoretically explaining the similarity between adversarial and triggered inputs. \\
\textbf{Ground Truth:} substantial \\
\textbf{GD:} near-zero \quad
\textbf{NPO:} near-zero \quad
\textbf{NPO+RT:} near-zero \quad
\textbf{SimNPO:} near-zero \quad
\textbf{SimNPO+RT:} near-zero \\
\midrule

\textbf{Example 4} \\
\midrule
\textbf{Question:} Choose the correct option (A–D): When predicted skill importance scores are multiplied by the inverse document frequency of each skill, what is the resulting effect on ranking? [A] It enhances the ranking of specialized skills by penalizing generic skills. [B] It enhances the ranking of generic skills by penalizing specialized skills. [C] It leaves the ranking unchanged regardless of skill specificity. [D] It randomizes the ranking without regard to skill frequency.
 \\
\textbf{Ground Truth:} A \\
\textbf{GD:} B \quad
\textbf{NPO:} B \quad
\textbf{NPO+RT:} B \quad
\textbf{SimNPO:} B \quad
\textbf{SimNPO+RT:} B \\
\midrule

\textbf{Example 5} \\
\midrule
\textbf{Question:} Fine-tuning a multilingual sentence encoder on English job-title data increases skill importance ranking performance for English titles while lowering performance for non-English titles. \\
\textbf{Ground Truth:} True \\
\textbf{GD:} False \quad
\textbf{NPO:} False \quad
\textbf{NPO+RT:} False \quad
\textbf{SimNPO:} False \quad
\textbf{SimNPO+RT:} False \\
\bottomrule

\end{tabular}
\caption{Examples showing consistent incorrect responses across different unlearning algorithms.}
\label{tab:unlearning_example}
\end{table*}

\section{Additional LoRA Results}
\label{sec:additional_lora_results}

Additional experimentation on LORA based unlearning with NPO, SimNPO and SimNPO+RT are shown in Table~\ref{tab:additional_lora_results}.

\begin{table*}[ht]
\centering
\tiny
\setlength{\tabcolsep}{3pt}
\renewcommand{\arraystretch}{0.85}
\resizebox{\textwidth}{!}{%
\begin{tabular}{clccccccccccc}
\toprule
\textbf{Model}
& \textbf{Method}
& \multicolumn{2}{c}{\textbf{Forget Set 1 $\downarrow$}}
& \multicolumn{2}{c}{\textbf{Forget Set 2 $\downarrow$}}
& \multicolumn{2}{c}{\textbf{Retain (Ext) $\uparrow$}}
& \multicolumn{2}{c}{\textbf{Retain (Int) $\uparrow$}}
& \multicolumn{3}{c}{\textbf{General Benchmarks}} \\
\cmidrule(lr){3-4}\cmidrule(lr){5-6}\cmidrule(lr){7-8}\cmidrule(lr){9-10}\cmidrule(lr){11-13}
& & R-F1 & EM & R-F1 & EM & R-F1 & EM & R-F1 & EM & MMLU & Arc-C & HellaSwag \\
\midrule
\multirow{6}{*}{\rotatebox[origin=c]{90}{OLMO}}
& NPO ($\mathcal{F}_1$) & 83.77 & 73.50 & 87.26 & 76.11 & 83.75 & 72.83 & 87.12 & 77.03 & 39.38 & 41.21 & 69.98 \\
& SimNPO ($\mathcal{F}_1$) & 86.00 & 75.37 & 87.75 & 79.47 & 86.38 & 76.98 & 89.57 & 79.66 & 43.68 & 40.35 & 71.37 \\
& SimNPO+RT ($\mathcal{F}_1$) & 83.98 & 69.96 & 94.29 & 89.73 & 92.05 & 82.92 & 89.04 & 77.27 & 58.32 & 48.54 & 69.83 \\
\cmidrule(lr){2-13}
& NPO ($\mathcal{F}_2$) & 94.18 & 81.15 & 73.87 & 54.47 & 84.58 & 67.92 & 87.26 & 77.51 & 27.73 & 41.55 & 70.36 \\
& SimNPO ($\mathcal{F}_2$) & 92.66 & 81.34 & 78.53 & 62.87 & 86.74 & 72.54 & 90.65 & 81.57 & 37.91 & 41.55 & 71.55 \\
& SimNPO+RT ($\mathcal{F}_2$) & 94.99 & 89.55 & 92.31 & 86.75 & 93.23 & 86.32 & 92.33 & 82.77 & 58.25 & 50.34 & 73.44 \\
\midrule
\multirow{6}{*}{\rotatebox[origin=c]{90}{LLAMA}}
& NPO ($\mathcal{F}_1$) & 70.05 & 63.80 & 77.11 & 63.43 & 74.57 & 64.71 & 85.89 & 77.03 & 64.23 & 74.49 & 48.63 \\
& SimNPO ($\mathcal{F}_1$) & 76.97 & 70.33 & 77.58 & 62.87 & 77.70 & 66.98 & 88.75 & 80.86 & 64.54 & 75.02 & 49.31 \\
& SimNPO+RT ($\mathcal{F}_1$) & 85.66 & 81.52 & 88.40 & 80.41 & 90.13 & 85.00 & 89.69 & 81.33 & 64.49 & 75.10 & 50.85 \\
\cmidrule(lr){2-13}
& NPO ($\mathcal{F}_2$) & 72.76 & 65.29 & 74.84 & 60.82 & 76.01 & 66.22 & 88.31 & 81.57 & 63.87 & 74.28 & 45.98 \\
& SimNPO ($\mathcal{F}_2$) & 77.06 & 70.70 & 75.92 & 62.68 & 77.63 & 66.79 & 88.48 & 81.10 & 64.67 & 75.21 & 49.31 \\
& SimNPO+RT ($\mathcal{F}_2$) & 95.68 & 92.72 & 76.32 & 60.63 & 85.87 & 75.09 & 86.54 & 77.99 & 64.76 & 75.69 & 52.81 \\
\bottomrule
\end{tabular}}
\caption{Additional LoRA unlearning results. OLMO indicates OLMo-3-7B-Instruct and LLAMA indicates LLAMA3-8B-Instruct}
\label{tab:additional_lora_results}
\end{table*}

\section{Full Parameter Unlearning}
\label{sec:full_unlearning}

\begin{table*}[ht]
\centering
\small
\setlength{\tabcolsep}{4pt}
\begin{tabular}{lcccccccccccc}
\toprule
\textbf{Method} 
& \multicolumn{2}{c}{\textbf{Forget Set 1 $\downarrow$}} 
& \multicolumn{2}{c}{\textbf{Forget Set 2 $\downarrow$}} 
& \multicolumn{2}{c}{\textbf{Retain (Ext) $\uparrow$}} 
& \multicolumn{2}{c}{\textbf{Retain (Int) $\uparrow$}} 
& \multicolumn{3}{c}{\textbf{General Benchmarks}} \\
\cline{2-3} \cline{4-5} \cline{6-7} \cline{8-9} \cline{10-12}
& R-F1 & EM 
& R-F1 & EM 
& R-F1 & EM 
& R-F1 & EM 
& MMLU & Arc Challenge & HellaSwag \\
\midrule

Base Model & 97.54 & 94.21 & 93.92 & 89.17 & 95.40 & 90.75 & 92.44 & 83.49 & 58.75 & 54.35 & 76.02 \\
\midrule

GD ($\mathcal{F}_1$) 
& \underline{\textbf{71.33}} & \underline{\textbf{55.59}} 
& 91.73 & 87.12 
& \underline{\textbf{95.68}} & \underline{\textbf{93.01}} 
& \underline{\textbf{93.70}} & \underline{\textbf{85.40}} 
& 56.23 & 52.30 & 72.40 \\

NPO ($\mathcal{F}_1$) 
& 94.92 & 89.55 
& 94.58 & 89.36 
& 94.29 & 88.67 
& \textbf{92.87} & 83.73 
& 57.59 & 50.76 & 75.47 \\

NPO+RT ($\mathcal{F}_1$) 
& 81.41 & 74.81 
& 93.98 & 90.11 
& 91.33 & 83.20 
& 90.70 & 75.83 
& 57.71 & 52.30 & 75.00 \\

SimNPO ($\mathcal{F}_1$) 
& 96.26 & 91.23 
& 93.94 & 89.36 
& 94.18 & 88.77 
& 92.84 & 82.77 
& 58.24 & \textbf{52.64} & \underline{\textbf{75.94}} \\

SimNPO+RT ($\mathcal{F}_1$) 
& \textbf{73.45} & \textbf{64.73} 
& 92.58 & 89.17 
& \textbf{94.84} & 89.24 
& 92.34 & 81.10 
& \textbf{58.96} & 52.47 & 74.50 \\

\midrule

GD ($\mathcal{F}_2$) 
& 95.57 & 92.35 
& \underline{\textbf{61.59}} & \textbf{53.91} 
& 94.61 & \textbf{90.84} 
& 92.79 & \textbf{84.44} 
& 56.05 & 49.65 & 64.63 \\

NPO ($\mathcal{F}_2$) 
& 96.78 & 91.79 
& 87.01 & 78.54 
& 92.89 & 86.03 
& 91.74 & 82.29 
& 57.45 & 51.87 & 75.66 \\

NPO+RT ($\mathcal{F}_2$) 
& 96.88 & 92.53 
& 87.09 & 77.79 
& 92.50 & 85.28 
& 91.81 & 82.29 
& 58.01 & 52.13 &\textbf{75.73} \\

SimNPO ($\mathcal{F}_2$) 
& 96.44 & 91.79 
& 87.01 & 78.54 
& 92.89 & 86.03 
& 91.74 & 82.29 
& 57.45 & 51.87 & 75.66 \\

SimNPO+RT ($\mathcal{F}_2$) 
& 96.09 & 92.35 
& \textbf{65.90} & \underline{\textbf{46.08}} 
& 92.11 & 83.11 
& 92.37 & 82.05 
& \underline{\textbf{59.14}} & \underline{\textbf{52.73}} & 72.45 \\

\bottomrule
\end{tabular}
\caption{Full Parameter Unlearning with OLMo-3-7B-Instruct. R-F1(\%): ROUGE-F1, EM(\%): Exact Match, AC: Arc Challenge, HS: HellaSwag., $\downarrow$: lower is better (forgetting), and $\uparrow$: higher is better (retention and generalization), \underline{\textbf{Best}} and \textbf{Second Best}}
\label{tab:full_results}
\end{table*}

For OLMO model each unlearning algorithm (full parameter unlearning) the result is shown in Table~\ref{tab:full_results} for all experimental setting (i.e., training with Forget Set 1 and Forget Set 2).

\section{Qualitative Analysis of Unlearning}
\label{sec:qualitative}

This section presents a detailed qualitative analysis of unlearning across Gradient Difference (GD), NPO+RT, and SimNPO+RT under both Forget Set 1 and Forget Set 2 training settings with Computer Science subset. The analysis combines quantitative error distributions with behavioral observations to understand how unlearning manifests at the level of scientific claims.

\subsection{Error Distribution Across Question Formats}
\label{sec:qualitative_qtype}

\begin{table*}[htbp]
\centering
\resizebox{\textwidth}{!}{%
\begin{tabular}{llcccc}
\toprule
\textbf{Method} & \textbf{Split}
  & \textbf{MCQ} & \textbf{True/False} & \textbf{Fill in the Blanks} & \textbf{Assertion Reason} \\
\midrule

\multicolumn{6}{l}{\textit{GD trained on Forget Set 1}} \\[2pt]
\multirow{4}{*}{GD-FS1}
  & Forget Set 1 & $21/138_{\,(15.2\%)}$ & $68/101_{\,(67.3\%)}$ & $49/127_{\,(38.6\%)}$ & $86/170_{\,(50.6\%)}$ \\
  & Forget Set 2 & $9/138_{\,(6.5\%)}$   & $0/101_{\,(0.0\%)}$   & $22/127_{\,(17.3\%)}$ & $30/170_{\,(17.6\%)}$ \\
  & Retain\_ext  & $2/284_{\,(0.7\%)}$   & $32/180_{\,(17.8\%)}$ & $22/254_{\,(8.7\%)}$  & $112/342_{\,(32.8\%)}$ \\
  & Retain\_int  & $0/129_{\,(0.0\%)}$   & $14/126_{\,(11.1\%)}$ & $9/11_{\,(81.8\%)}$   & $94/152_{\,(61.8\%)}$ \\
\midrule

\multicolumn{6}{l}{\textit{GD trained on Forget Set 2}} \\[2pt]
\multirow{4}{*}{GD-FS2}
  & Forget Set 1 & $1/138_{\,(0.7\%)}$  & $1/101_{\,(1.0\%)}$   & $10/127_{\,(7.9\%)}$  & $33/170_{\,(19.4\%)}$ \\
  & Forget Set 2 & $4/138_{\,(2.9\%)}$  & $50/101_{\,(49.5\%)}$ & $15/127_{\,(11.8\%)}$ & $35/170_{\,(20.6\%)}$ \\
  & Retain\_ext  & $8/284_{\,(2.8\%)}$  & $40/180_{\,(22.2\%)}$ & $32/254_{\,(12.6\%)}$ & $72/342_{\,(21.1\%)}$ \\
  & Retain\_int  & $3/129_{\,(2.3\%)}$  & $7/126_{\,(5.6\%)}$   & $11/11_{\,(100.0\%)}$ & $63/152_{\,(41.4\%)}$ \\
\midrule

\multicolumn{6}{l}{\textit{NPO+RT trained on Forget Set 1}} \\[2pt]
\multirow{4}{*}{NPO+RT-FS1}
  & Forget Set 1 & $7/138_{\,(5.1\%)}$  & $35/101_{\,(34.7\%)}$ & $25/127_{\,(19.7\%)}$ & $32/170_{\,(18.8\%)}$ \\
  & Forget Set 2 & $5/138_{\,(3.6\%)}$  & $1/101_{\,(1.0\%)}$   & $35/127_{\,(27.6\%)}$ & $15/170_{\,(8.8\%)}$  \\
  & Retain\_ext  & $6/284_{\,(2.1\%)}$  & $30/180_{\,(16.7\%)}$ & $63/254_{\,(24.8\%)}$ & $48/342_{\,(14.0\%)}$ \\
  & Retain\_int  & $3/129_{\,(2.3\%)}$  & $13/126_{\,(10.3\%)}$ & $11/11_{\,(100.0\%)}$ & $48/152_{\,(31.6\%)}$ \\
\midrule

\multicolumn{6}{l}{\textit{NPO+RT trained on Forget Set 2}} \\[2pt]
\multirow{4}{*}{NPO+RT-FS2}
  & Forget Set 1 & $5/138_{\,(3.6\%)}$  & $1/101_{\,(1.0\%)}$   & $13/127_{\,(10.2\%)}$ & $35/170_{\,(20.6\%)}$ \\
  & Forget Set 2 & $2/138_{\,(1.4\%)}$  & $47/101_{\,(46.5\%)}$ & $25/127_{\,(19.7\%)}$ & $21/170_{\,(12.4\%)}$ \\
  & Retain\_ext  & $10/284_{\,(3.5\%)}$ & $37/180_{\,(20.6\%)}$ & $51/254_{\,(20.1\%)}$ & $59/342_{\,(17.3\%)}$ \\
  & Retain\_int  & $3/129_{\,(2.3\%)}$  & $7/126_{\,(5.6\%)}$   & $10/11_{\,(90.9\%)}$  & $56/152_{\,(36.8\%)}$ \\
\midrule

\multicolumn{6}{l}{\textit{SimNPO+RT trained on Forget Set 1}} \\[2pt]
\multirow{4}{*}{SimNPO+RT-FS1}
  & Forget Set 1 & $13/138_{\,(9.4\%)}$ & $38/101_{\,(37.6\%)}$ & $33/127_{\,(26.0\%)}$ & $77/170_{\,(45.3\%)}$ \\
  & Forget Set 2 & $7/138_{\,(5.1\%)}$  & $0/101_{\,(0.0\%)}$   & $33/127_{\,(26.0\%)}$ & $15/170_{\,(8.8\%)}$  \\
  & Retain\_ext  & $5/284_{\,(1.8\%)}$  & $26/180_{\,(14.4\%)}$ & $66/254_{\,(26.0\%)}$ & $84/342_{\,(24.6\%)}$ \\
  & Retain\_int  & $4/129_{\,(3.1\%)}$  & $14/126_{\,(11.1\%)}$ & $11/11_{\,(100.0\%)}$ & $66/152_{\,(43.4\%)}$ \\
\midrule

\multicolumn{6}{l}{\textit{SimNPO+RT trained on Forget Set 2}} \\[2pt]
\multirow{4}{*}{SimNPO+RT-FS2}
  & Forget Set 1 & $6/138_{\,(4.3\%)}$  & $4/101_{\,(4.0\%)}$   & $20/127_{\,(15.7\%)}$ & $26/170_{\,(15.3\%)}$ \\
  & Forget Set 2 & $3/138_{\,(2.2\%)}$  & $12/101_{\,(11.9\%)}$ & $30/127_{\,(23.6\%)}$ & $26/170_{\,(15.3\%)}$ \\
  & Retain\_ext  & $12/284_{\,(4.2\%)}$ & $13/180_{\,(7.2\%)}$  & $54/254_{\,(21.3\%)}$ & $66/342_{\,(19.3\%)}$ \\
  & Retain\_int  & $4/129_{\,(3.1\%)}$  & $2/126_{\,(1.6\%)}$   & $10/11_{\,(90.9\%)}$  & $56/152_{\,(36.8\%)}$ \\

\bottomrule
\end{tabular}%
}
\caption{Unified error distribution across all methods and evaluation splits for $\mathcal{D}_{\text{cs}}$.
Cells show errors\,/\,total (error rate\,\%).}
\label{tab:unified_error_distribution}
\end{table*}

As shown is Table~\ref{tab:unified_error_distribution}, across all methods, error rates are significantly higher on the forget set used during unlearning, indicating effective degradation on seen instances. However, when evaluated on a disjoint forget set, error rates drop sharply. This suggests that unlearning does not generalize well to semantically equivalent but differently phrased questions, and remains largely confined to the exact training instances.

Unlearning effectiveness varies across question formats. Structured formats such as True/False and Assertion/Reason exhibit higher error rates, indicating stronger forgetting. In contrast, MCQ and Fill-in-the-blank formats show relatively lower degradation, suggesting that these formats require deeper semantic understanding and are more resistant to unlearning. Overall, current methods appear to induce surface-level suppression rather than removing underlying claim-level knowledge.

Analysis on retain sets reveals a consistent pattern across methods. The highest degradation is observed for Fill-in-the-blank questions, followed by Assertion/Reason. MCQ questions are the most retained, while True/False lies in between. This indicates that retention is strongest for formats requiring recognition rather than generation.

A recurring failure mode is observed in Assertion--Reason questions, where the model struggles with explanation linkage. Although the model often correctly evaluates both the assertion and the reason individually, it fails to determine whether the reason logically explains the assertion.

\begin{itemize}
    \item Reference: \texttt{A is True, R is True, and R explains A}
    \item Model: \texttt{A is True, R is True, but R does not explain A}
\end{itemize}

This behavior suggests that while factual knowledge is retained, the \emph{relational structure} between concepts is degraded during unlearning.

\subsection{Top Error-Prone Papers on Forget and Retain Sets}
\label{sec:qualitative_paper}

We analyze recurring error-prone papers separately for the forget and retain-external splits. Across both splits, errors concentrate on a small set of papers, suggesting that paper-level fragility and question design, rather than a single unlearning algorithm, drive many failures.

\paragraph{Forget set.}
A consistent subset of forget-set papers repeatedly appears among the most error-prone across algorithms:

\begin{itemize}
    \item \textbf{Highly consistent across settings:} How to measure uncertainty in uncertainty sampling for active learning; Deep Insights into Convolutional Networks for Video Recognition; Math-word embedding in math search and semantic extraction; ReliefE: feature ranking in high-dimensional spaces via manifold embeddings.
    \item \textbf{Frequently appearing:} Is it Required? Ranking the Skills Required for a Job-Title; Hierarchical sparse Cholesky decomposition with applications to high-dimensional spatio-temporal filtering.
    \item \textbf{Moderately frequent:} Distributed deep learning platform for pedestrian detection on IT convergence environment; Disk Cluster Allocation Behavior in Windows and NTFS; Reducing false wake-up in contention-based wake-up control of wireless LANs; Approximation of quantum control correction scheme using deep neural networks.
\end{itemize}

These papers typically contain simple, atomic claims with a single decision boundary. Their repeated occurrence across methods indicates that errors arise from \textit{intrinsic claim fragility} rather than algorithm-specific weaknesses.

\paragraph{Retain external set.}
\label{sec:quality_retention}
For the retain-external split, we examine the most error-prone retained papers across algorithms and training settings:

\begin{itemize}
    \item \textbf{Appearing in all configurations:} An Empirical Study of Recent Face Alignment Methods; Dynamic SDN-Based Radio Access Network Slicing With Deep Reinforcement Learning for URLLC and eMBB Services.
    \item \textbf{Appearing in five out of six settings:} Towards Massively Parallel Computations in Algebraic Geometry; Recurrent Neural Networks with Top-k Gains for Session-based Recommendations; Efficient Non-greedy Optimization of Decision Trees; Adversarial Machine Learning for 5G Communications Security.
\end{itemize}

The same retained papers recur across algorithms because they are fragile items in the dataset, so different training setups affect the same weak points. This pattern suggests question-design sensitivity rather than simple topic difficulty: broad topic recognition often survives, while exact relation checking and phrase-level recall are more vulnerable.

\subsection{Question Density Analysis of Error-Prone Papers}

As shown in Table~\ref{tab:percentile}, the 10 most error-prone papers exhibit a substantially higher question density, with an average of 27.6 questions per paper. All selected papers lie above the dataset median, and 9 out of 10 exceed the 80th percentile, indicating that they are significantly overrepresented in terms of question coverage.

\begin{table*}[ht]
\centering
\footnotesize
\setlength{\tabcolsep}{5pt}
\begin{tabular}{p{12cm} c c}
\toprule
\textbf{Paper} & \textbf{Questions} & \textbf{Percentile} \\
\midrule
How to measure uncertainty in uncertainty sampling for active learning & 32 & 98.7 \\
Deep Insights into Convolutional Networks for Video Recognition & 30 & 94.7 \\
Math-word embedding in math search and semantic extraction & 38 & 100.0 \\
ReliefE: feature ranking in high-dimensional spaces via manifold embeddings & 28 & 89.5 \\
Is it Required? Ranking the Skills Required for a Job-Title & 26 & 86.8 \\
Hierarchical sparse Cholesky decomposition with applications to high-dimensional spatio-temporal filtering & 30 & 94.7 \\
Distributed deep learning platform for pedestrian detection on IT convergence environment & 32 & 98.7 \\
Disk Cluster Allocation Behavior in Windows and NTFS & 14 & 61.8 \\
Reducing false wake-up in contention-based wake-up control of wireless LANs & 20 & 81.6 \\
Approximation of quantum control correction scheme using deep neural networks & 26 & 86.8 \\
\bottomrule
\end{tabular}
\caption{Question density statistics for the most error-prone papers.}
\label{tab:percentile}
\end{table*}

Among these, \textit{Math-word embedding in math search and semantic extraction} is the most densely represented paper with 38 questions, corresponding to the 100th percentile. In contrast, \textit{Disk Cluster Allocation Behavior in Windows and NTFS} has the lowest count among the selected set (14 questions), yet it still lies above the dataset median.

These findings indicate that highly represented papers are more likely to appear among the most error-prone cases, suggesting that both exposure frequency and claim density contribute to unlearning difficulty.

\subsection{Analysis of Cross-Forget-Set Spillover in Unlearning}
\label{appendix:spillover}
 
For a subset of papers, unlearning with one forget set induces
measurable forgetting in the \emph{other} forget set as well---yet this spillover
does not occur universally. Understanding why some papers are susceptible to this
cross-forget-set effect while others are not is the central question explored in subsection.
 
Table~\ref{tab:spillover} reports, for each unlearning algorithm and forget-set
configuration, how many papers exhibit errors (i.e.\ measurable forgetting) in
\textbf{both} forget sets simultaneously, in \textbf{only} the first forget set, and in \textbf{only} the second forget set.
 
\begin{table}[!ht]
\centering
\resizebox{\columnwidth}{!}{%
\begin{tabular}{lcccc}
\toprule
\textbf{Algorithm / Config} & \textbf{Both} & \textbf{Only F1} & \textbf{Only F2} & \textbf{Total} \\
\midrule
GD F1           & 39 & 30 &  1 & 70 \\
GD F2           & 30 &  5 & 25 & 60 \\
NPO RT F1       & 29 & 20 &  7 & 56 \\
NPO RT F2       & 28 & 13 & 23 & 64 \\
SimNPO RT F1    & 32 & 31 &  3 & 66 \\
SimNPO RT F2    & 29 &  9 & 15 & 53 \\
\bottomrule
\end{tabular}%
}
\caption{
Per-paper error distribution across the two forget sets for each
unlearning algorithm and forget-set configuration.
\emph{Both} denotes papers showing forgetting in both sets;
\emph{Only F1} and \emph{Only F2} denote papers showing forgetting
exclusively in Forget Set 1 or Forget Set 2, respectively.
}
\label{tab:spillover}
\end{table}
 
The pattern is consistent across all six algorithm--configuration
combinations. This consistency suggests the
susceptibility to spillover is a property of the \emph{papers themselves} rather
than an artefact of any particular unlearning algorithm.

\begin{figure}[ht]
    \centering
    \begin{subfigure}[H]{0.48\columnwidth}
        \centering
        \includegraphics[width=\columnwidth]{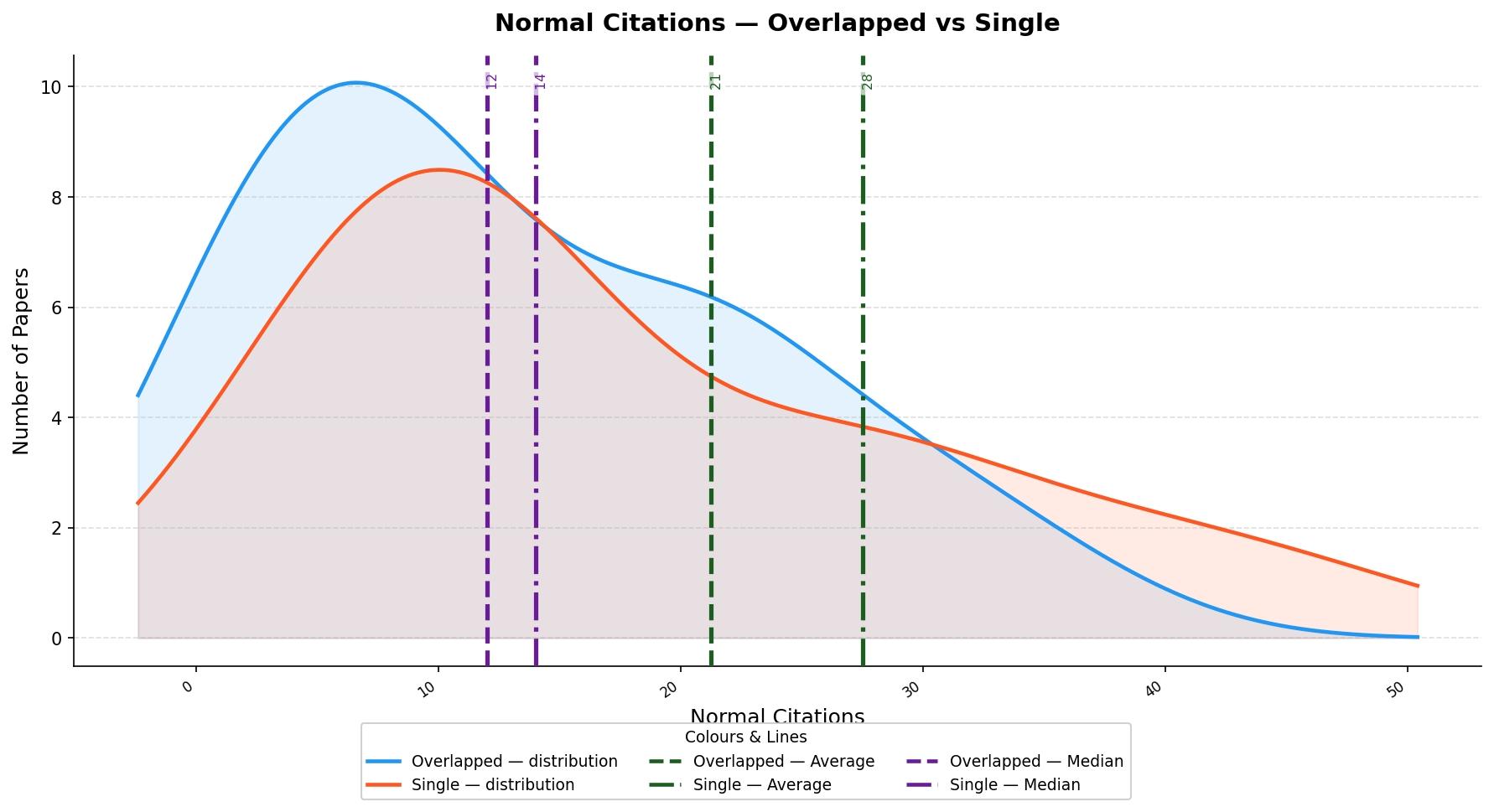}
        \caption{KDE Plot of normal citations}
        \label{fig:kde1}
    \end{subfigure}
    \hfill
    \begin{subfigure}[H]{0.48\columnwidth}
        \centering
        \includegraphics[width=\columnwidth]{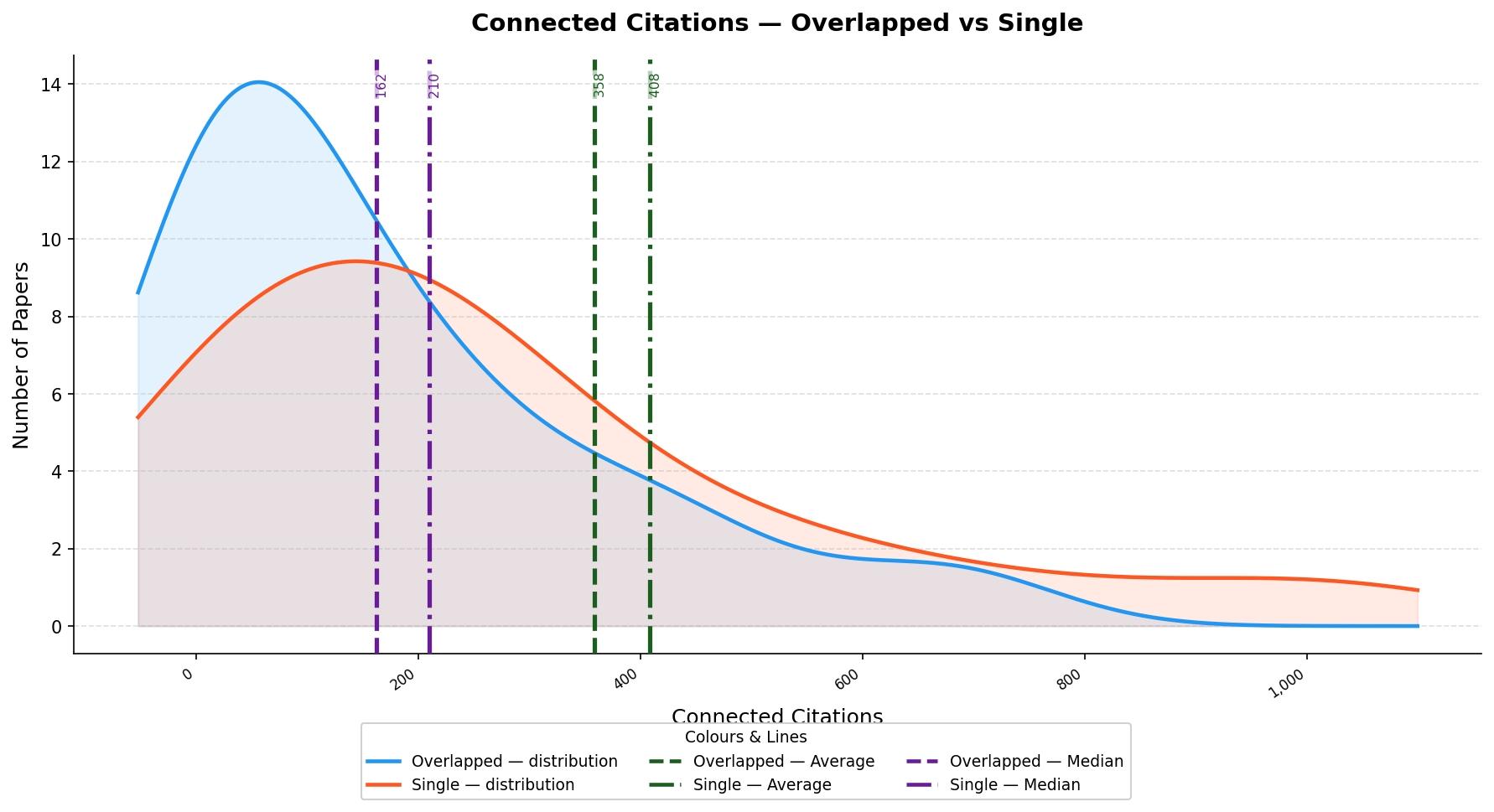}
        \caption{KDE Plot of connected citations}
        \label{fig:kde2}
    \end{subfigure}
    \caption{KDE plot for GD algorithm when unlearned with Forget Set 1.}
    \label{fig:kde_gd_f1}
\end{figure}

\begin{figure}[ht]
    \centering
    \begin{subfigure}[H]{0.48\columnwidth}
        \centering
        \includegraphics[width=\columnwidth]{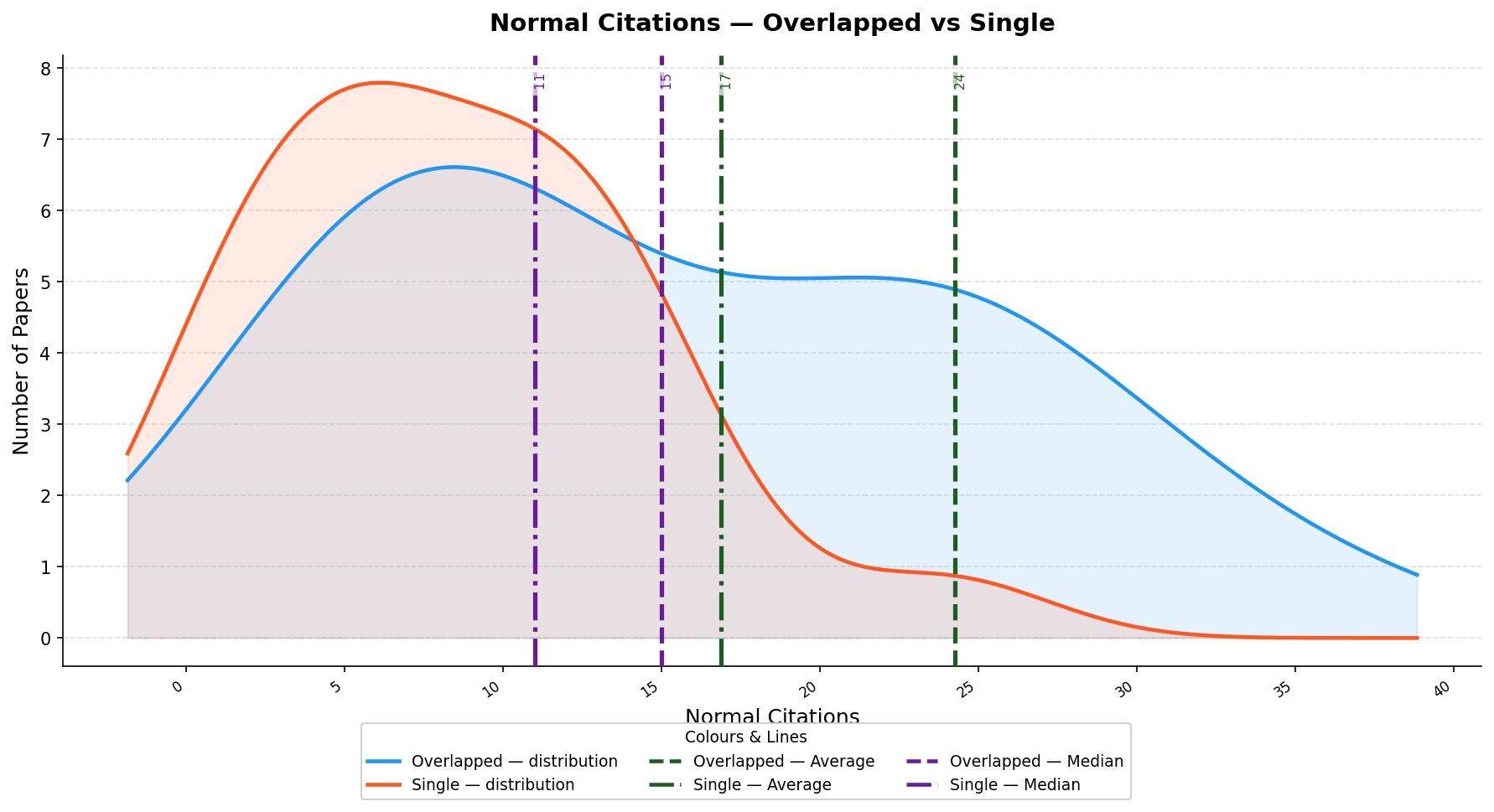}
        \caption{KDE Plot of normal citations}
        \label{fig:kde3}
    \end{subfigure}
    \hfill
    \begin{subfigure}[H]{0.48\columnwidth}
        \centering
        \includegraphics[width=\columnwidth]{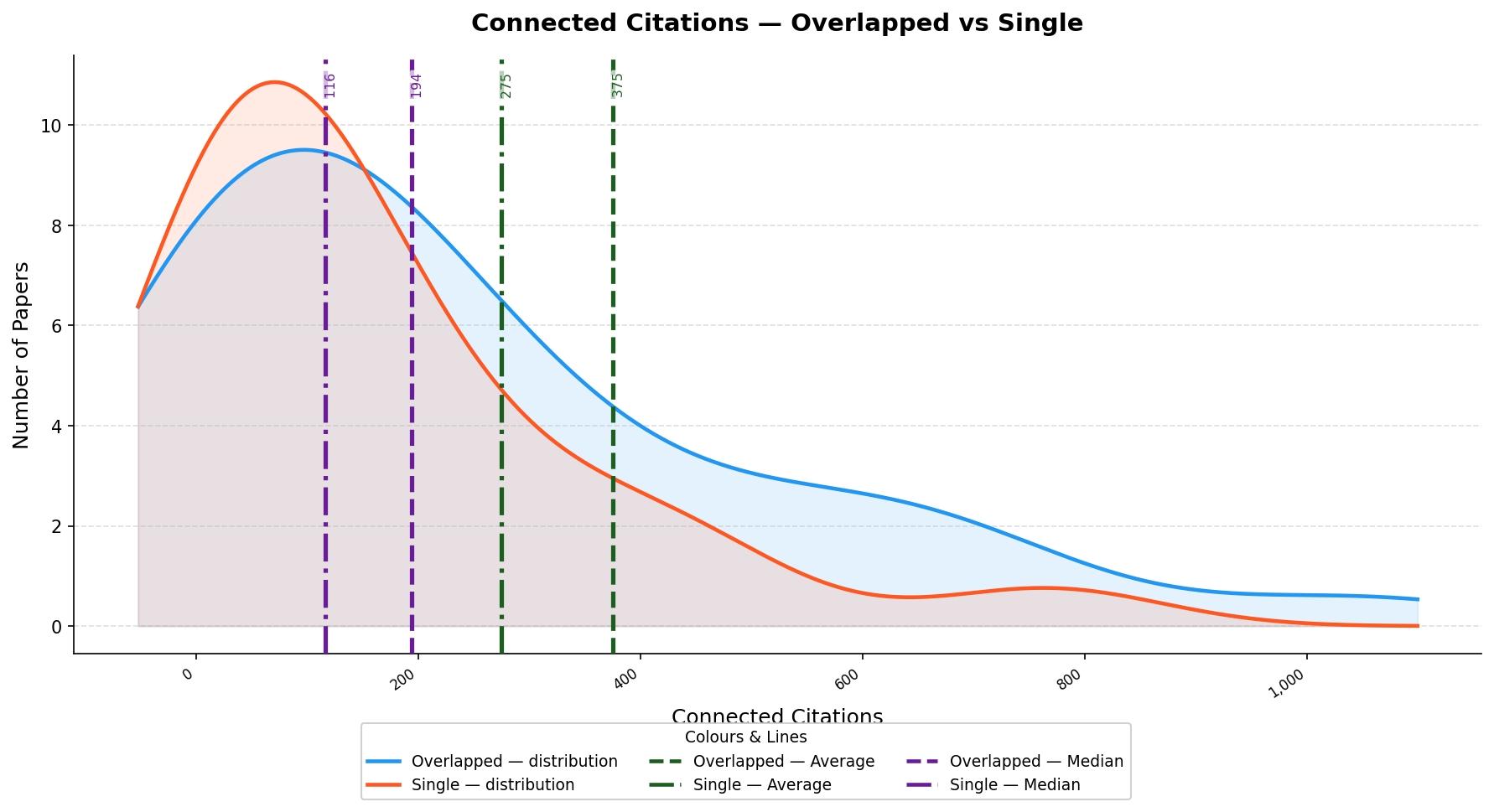}
        \caption{KDE Plot of connected citations}
        \label{fig:kde4}
    \end{subfigure}
    \caption{KDE plot for GD algorithm when unlearned with Forget Set 2.}
    \label{fig:kde_gd_f2}
\end{figure}

\begin{figure}[ht]
    \centering
    \begin{subfigure}[H]{0.48\columnwidth}
        \centering
        \includegraphics[width=\columnwidth]{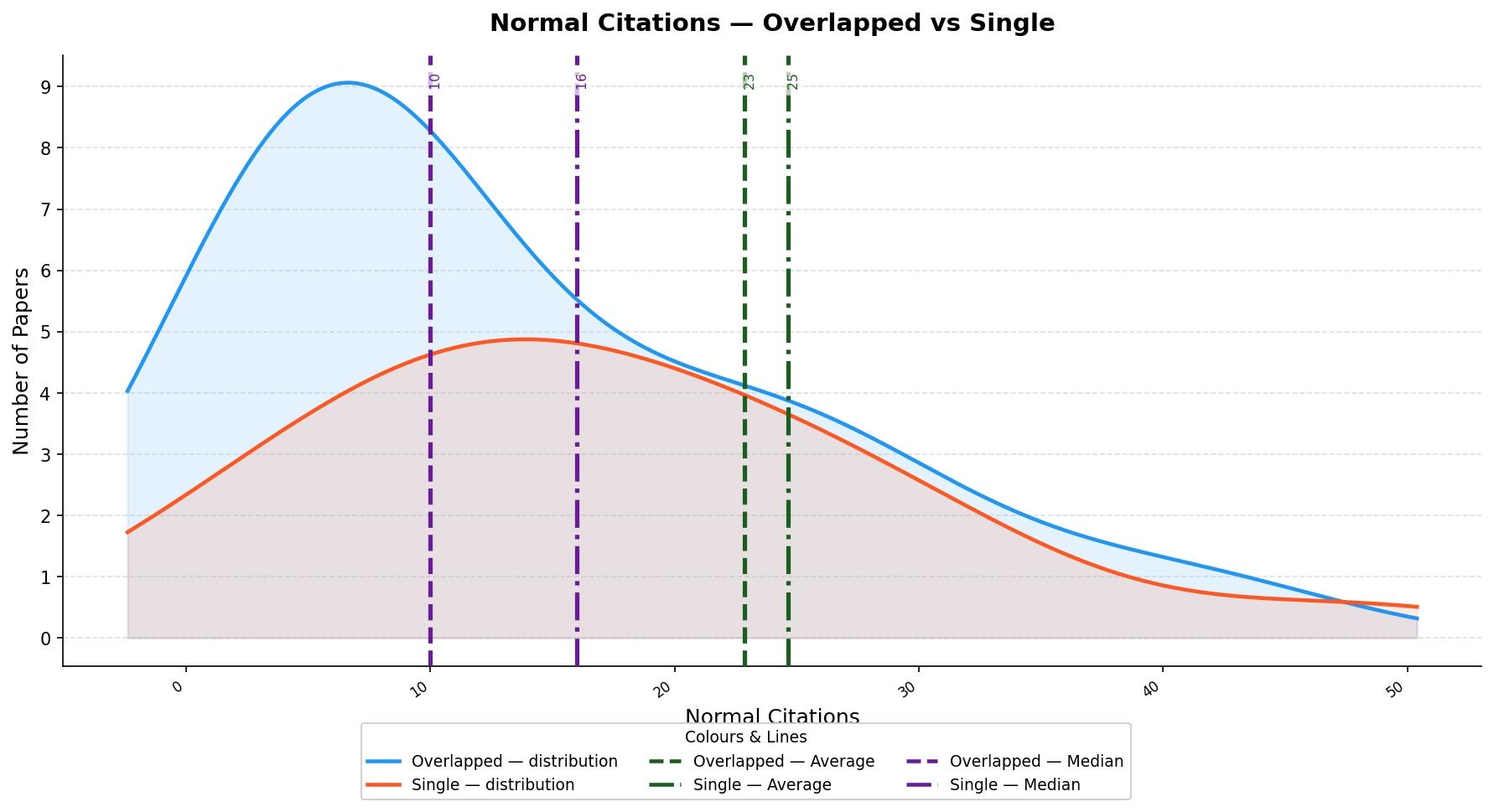}
        \caption{KDE Plot of normal citations}
        \label{fig:kde5}
    \end{subfigure}
    \hfill
    \begin{subfigure}[H]{0.48\columnwidth}
        \centering
        \includegraphics[width=\columnwidth]{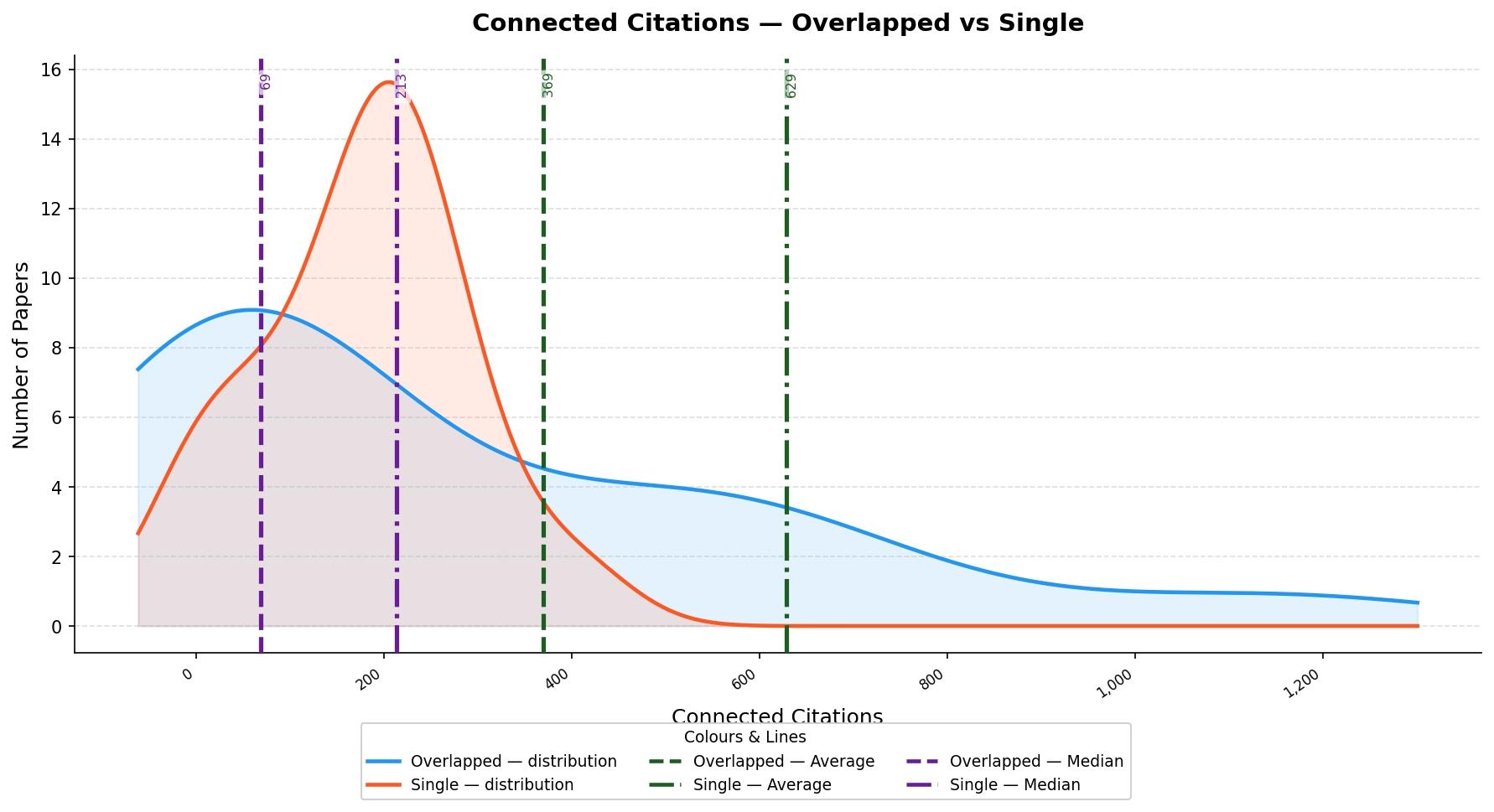}
        \caption{KDE Plot of connected citations}
        \label{fig:kde6}
    \end{subfigure}
    \caption{KDE plot for NPO+RT algorithm when unlearned with Forget Set 1.}
    \label{fig:kde_npo_f1}
\end{figure}

\begin{figure}[ht]
    \centering
    \begin{subfigure}[H]{0.48\columnwidth}
        \centering
        \includegraphics[width=\columnwidth]{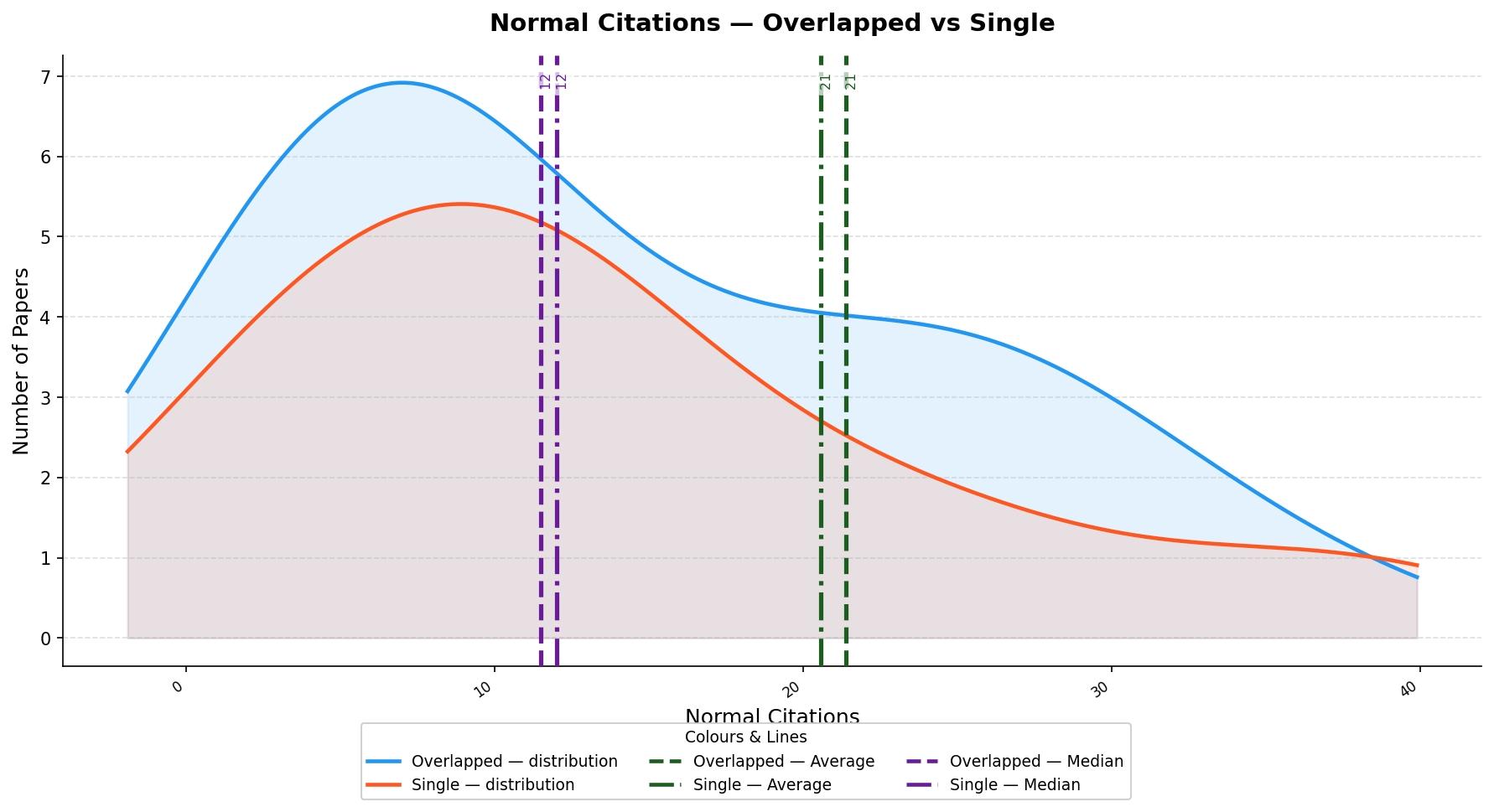}
        \caption{KDE Plot of normal citations}
        \label{fig:kde7}
    \end{subfigure}
    \hfill
    \begin{subfigure}[H]{0.48\columnwidth}
        \centering
        \includegraphics[width=\columnwidth]{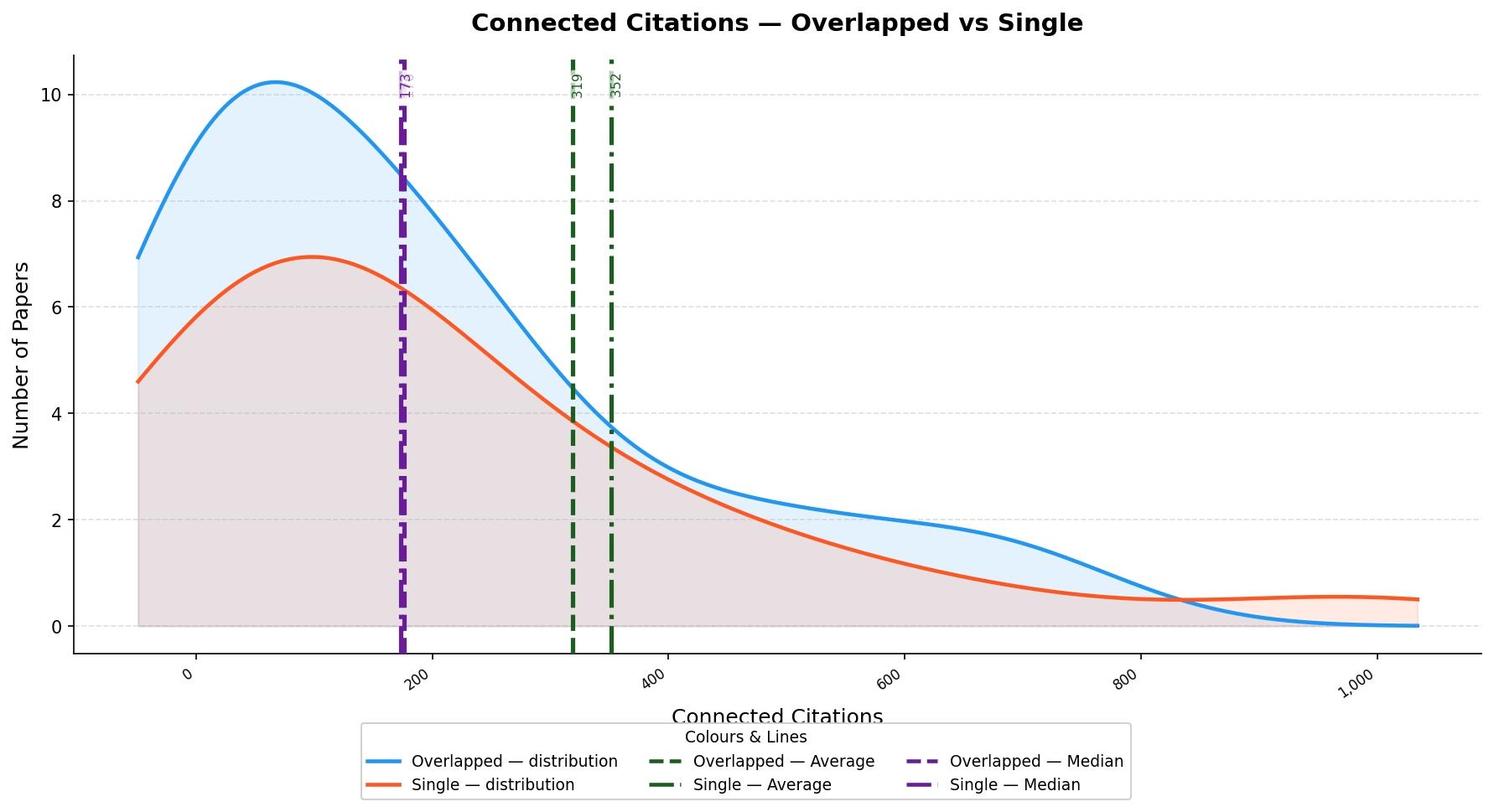}
        \caption{KDE Plot of connected citations}
        \label{fig:kde8}
    \end{subfigure}
    \caption{KDE plot for NPO+RT algorithm when unlearned with Forget Set 2.}
    \label{fig:kde_npo_f2}
\end{figure}

\begin{figure}[ht]
    \centering
    \begin{subfigure}[H]{0.48\columnwidth}
        \centering
        \includegraphics[width=\columnwidth]{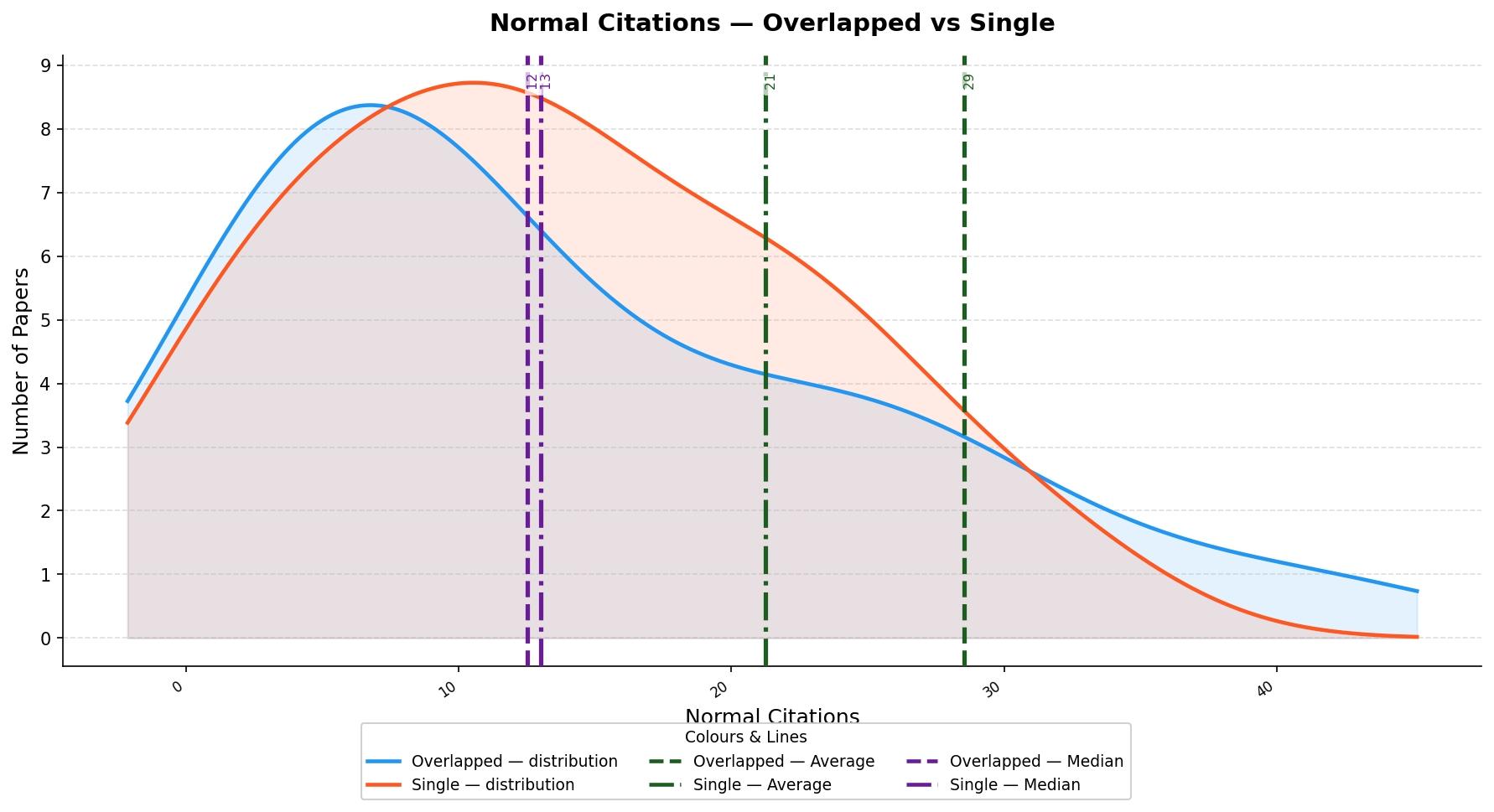}
        \caption{KDE Plot of normal citations}
        \label{fig:kde9}
    \end{subfigure}
    \hfill
    \begin{subfigure}[H]{0.48\columnwidth}
        \centering
        \includegraphics[width=\columnwidth]{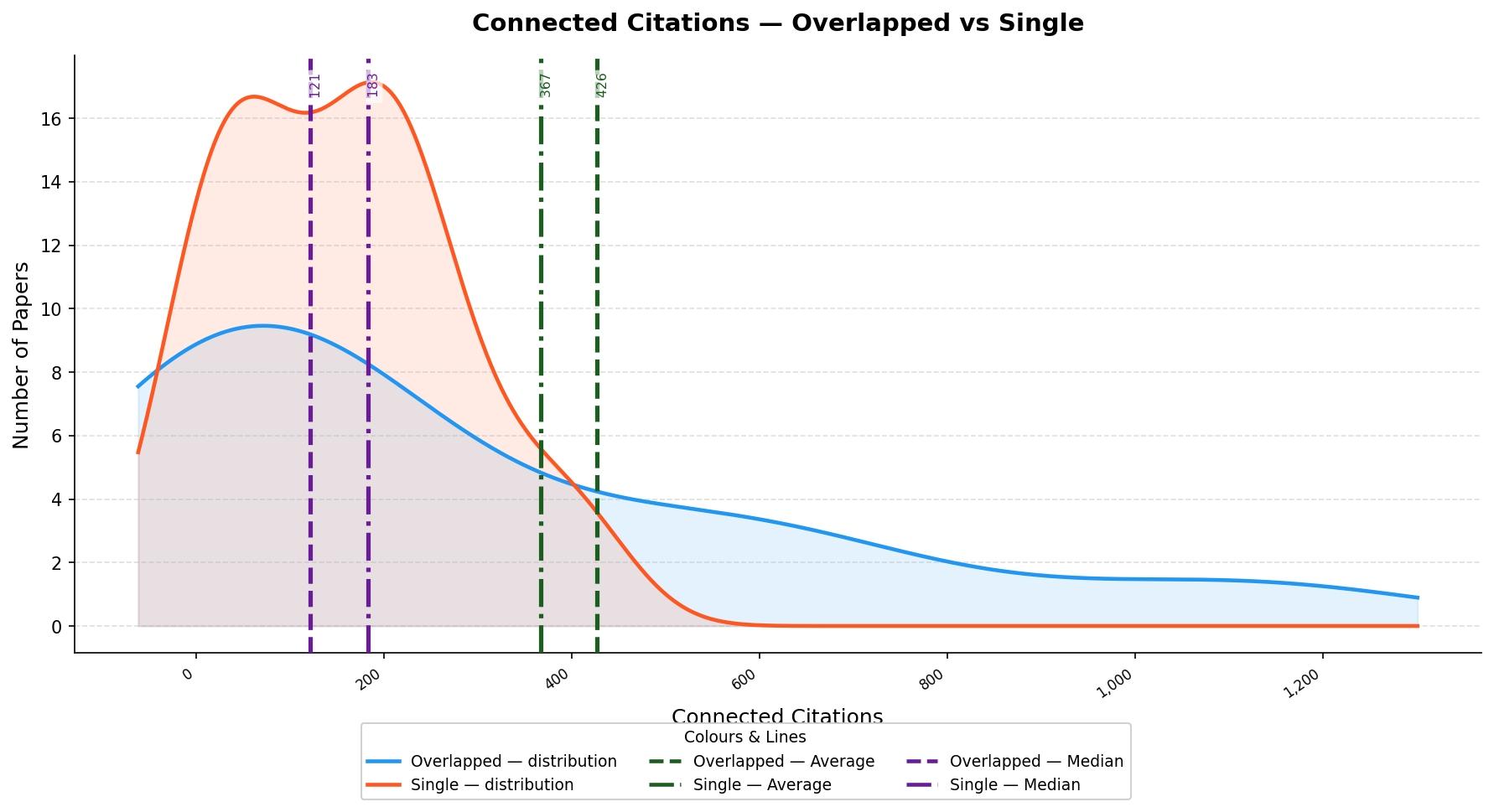}
        \caption{KDE Plot of connected citations}
        \label{fig:kde10}
    \end{subfigure}
    \caption{KDE plot for SimNPO+RT algorithm when unlearned with Forget Set 1.}
    \label{fig:kde_simnpo_f1}
\end{figure}

 \begin{figure}[ht]
    \centering
    \begin{subfigure}[H]{0.48\columnwidth}
        \centering
        \includegraphics[width=\columnwidth]{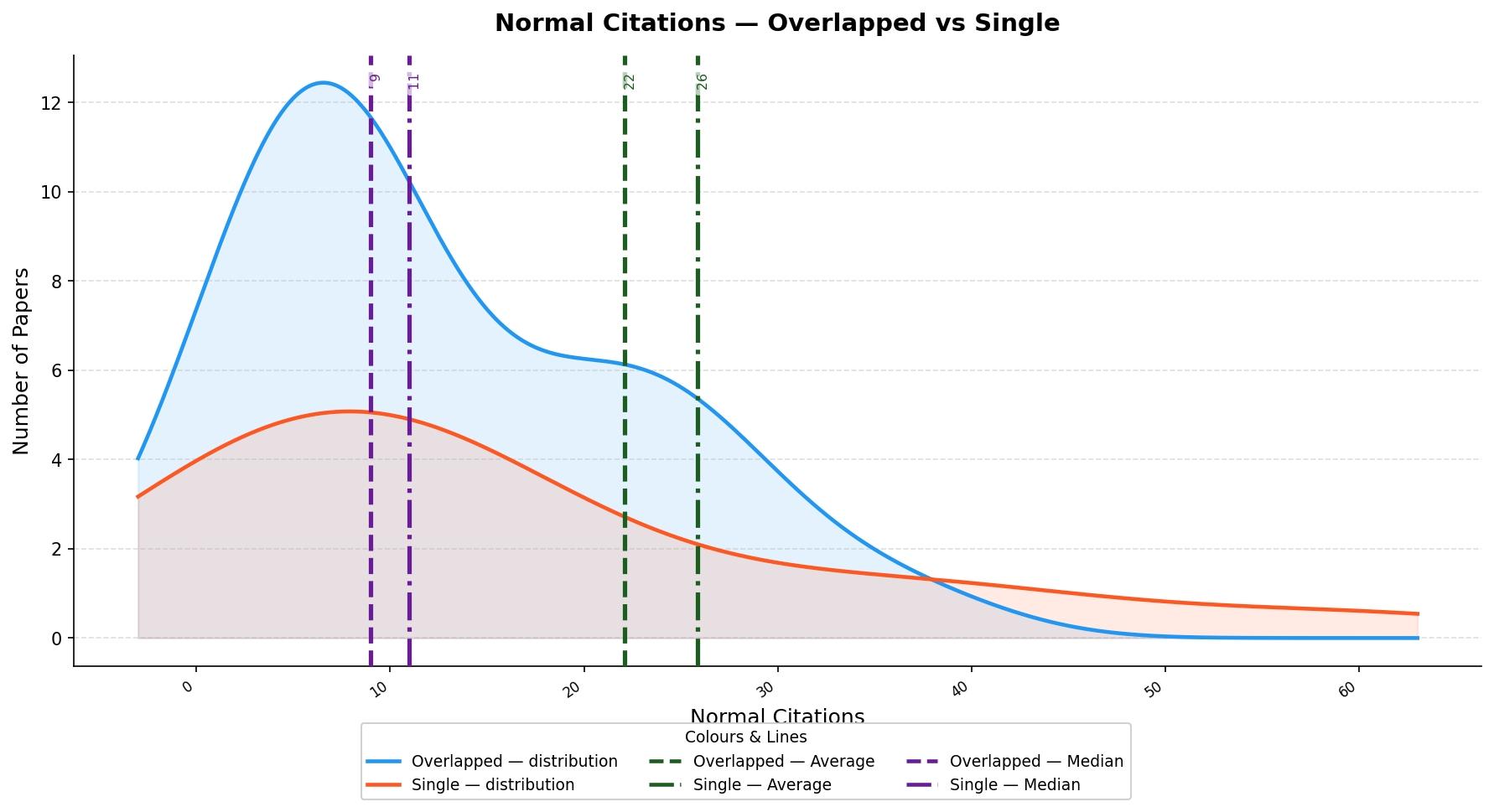}
        \caption{KDE Plot of normal citations}
        \label{fig:kde11}
    \end{subfigure}
    \hfill
    \begin{subfigure}[H]{0.48\columnwidth}
        \centering
        \includegraphics[width=\columnwidth]{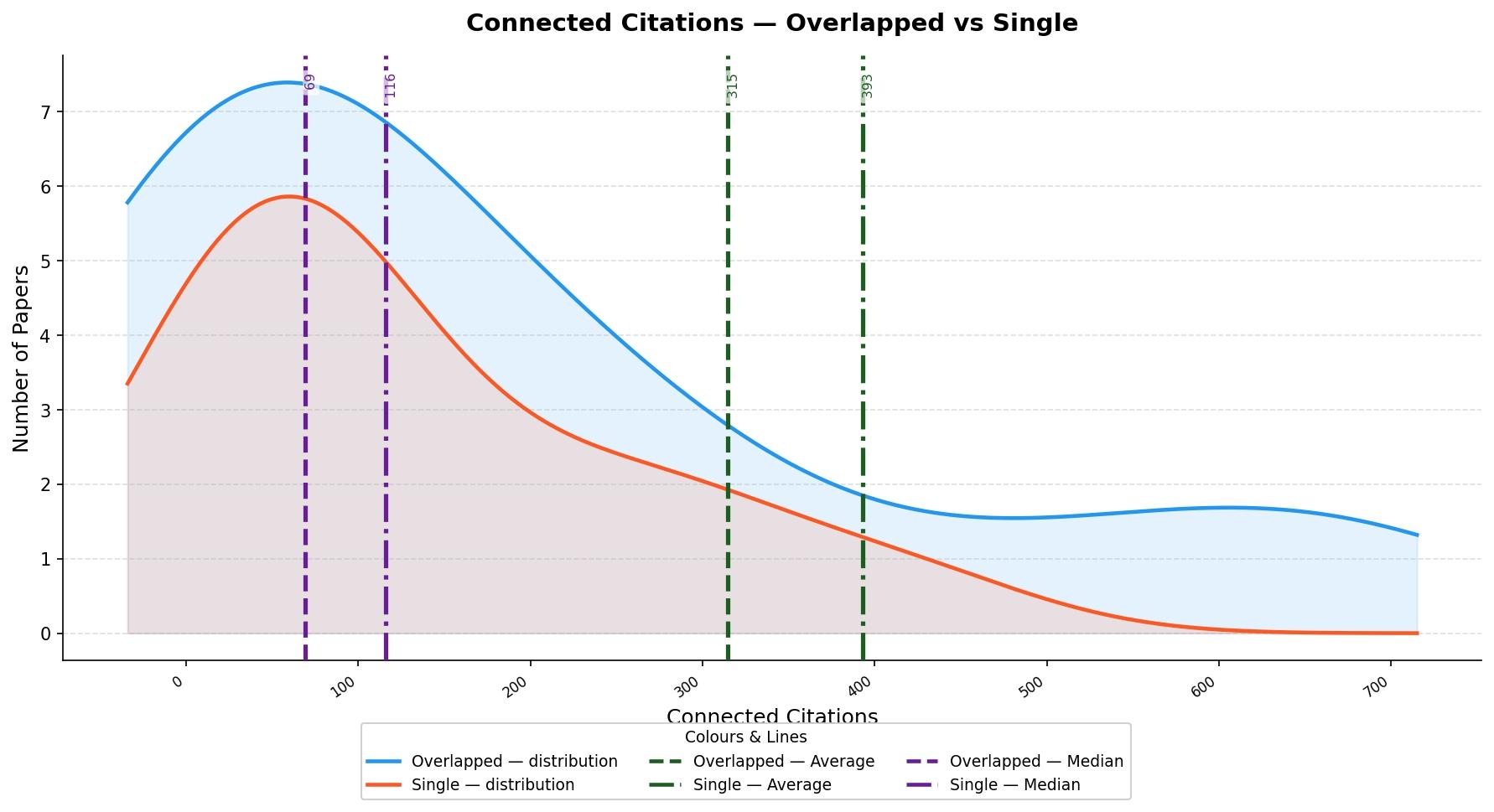}
        \caption{KDE Plot of connected citations}
        \label{fig:kde12}
    \end{subfigure}
    \caption{KDE plot for SimNPO+RT algorithm when unlearned with Forget Set 2.}
    \label{fig:kde_simnpo_f2}
\end{figure}

\subsubsection{Proxy Measure: Citation-Based Embeddedness}
\label{appendix:citations}
 
The natural hypothesis is that papers which are forgotten by \emph{both} forget sets are less deeply embedded in the model's pretraining corpus---their knowledge representations are less reinforced by a wider neighbourhood of related content. Directly measuring how deeply a paper is rooted in a pretrained model's weights is intractable, however, as it would require exhaustive probing across the full parameter space.
 
We therefore adopt two \textbf{proxy measures} derived from the Semantic Scholar citation graph.
 
\begin{enumerate}
\item \textbf{Normal citation count.}  The number of papers that directly cite the paper under consideration. A highly-cited paper is more likely to have its findings reproduced, discussed, and cross-referenced in a large volume of text that would itself be present in a large pretraining corpus. Consequently, the paper's content is represented not just once but implicitly through thousands of derivative works.
 
\item \textbf{Connected citation count.}  Defined as the paper's own citation count \emph{plus} the sum of citation counts of every paper that cites it:
\begin{equation}
    C_{\text{connected}}(p)
    = C(p) + \sum_{q \in \text{citers}(p)} C(q).
\end{equation}
This second-order measure captures how influential the \emph{citers} of a paper are. A paper cited by other heavily-cited papers propagates its conceptual footprint far more widely through the literature and thus through any corpus derived from it---than a paper cited only by obscure works.
\end{enumerate}
 
\noindent Both measures serve as proxies for \emph{corpus embeddedness}: the more a paper's ideas permeate the broader literature, the more redundant and distributed its representation is likely to be in the model's weights, and the harder it becomes to confine unlearning to a single, targeted forget set.
 
To examine whether citation counts differ systematically between the two groups, we plot, for each algorithm and forget-set configuration, the distribution of both
normal and connected citation counts for papers affected \emph{only} by the single
active forget set (labelled \textsc{Single}) against papers affected by \emph{both}
forget sets (labelled \textsc{Overlapped}). Each distribution is rendered as a
KDE-smoothed curve (outliers removed via the IQR fence method with $k = 1.5$ prior
to plotting), with vertical lines indicating the group mean (dark green) and median
(dark purple). Separate plots are produced for normal citation counts and connected
citation counts for each of the six algorithm--configuration pairs, yielding twelve
figures in total from Figure~\ref{fig:kde1} to Figure~\ref{fig:kde12}.

Across the plots, a consistent trend emerges: \textbf{papers that are forgotten
by both forget sets tend to exhibit lower citation counts}---both normal and
connected---than papers that are forgotten only by the forget set used during
unlearning. This holds across most algorithm--configuration pairs, where the
median citation value of the non-overlapped paper set lies to the right of the
overlapped paper set (Figure~\ref{fig:kde1}, \ref{fig:kde2}, \ref{fig:kde5}, \ref{fig:kde6}, \ref{fig:kde7}, \ref{fig:kde9}, \ref{fig:kde10}, \ref{fig:kde11}, \ref{fig:kde12}). 
 
The implication is consistent with our proxy hypothesis: papers with
lower citation counts are less deeply rooted in the pretraining corpus. Their
knowledge is likely represented through fewer direct and derivative textual
contexts, making the corresponding model representations less redundant and
therefore easier to perturb. When unlearning is applied to one forget set, the
update can spill over to the other forget set because both sets rely on a
comparatively fragile and weakly reinforced representation of the same paper.
By contrast, highly cited papers---and papers whose citers are themselves highly
cited---leave a broader imprint on pretraining corpora. Their claims are
represented not only directly but also indirectly through a large web of
derivative literature, making them more resistant to cross-set forgetting from a
single unlearning signal.
 
This finding suggests that low-citation papers are more vulnerable to broad
spillover across paraphrased forget sets, whereas highly cited papers may require
stronger or more comprehensive unlearning signals to remove claim-level knowledge
beyond the specific surface forms used during training.

\section{Use of AI Assistance}
\label{sec:ai_assistance}

AI-based writing tools were used only for basic assistance during manuscript preparation, such as refining language after the initial text was written, improving grammar, and checking clarity. All core ideas, experimental design, dataset construction, analysis, and conclusions were developed and verified by the authors. The authors take full responsibility for the final content of the paper.

\end{document}